\documentclass[lettersize,journal]{IEEEtran}
\usepackage{amsmath,amsfonts}
\usepackage{algorithmic}
\usepackage{algorithm}
\usepackage{array}
\usepackage[caption=false,font=normalsize,labelfont=sf,textfont=sf]{subfig}
\usepackage{textcomp}
\usepackage{stfloats}
\usepackage{url}
\usepackage{verbatim}
\usepackage{graphicx}
\usepackage{cite}
\usepackage{microtype}
\usepackage{inconsolata}
\usepackage{amssymb}
\usepackage{multirow}
\usepackage{multicol}
\usepackage{booktabs}
\usepackage{adjustbox}
\usepackage{enumitem}
\usepackage{bm}
\usepackage{comment}
\usepackage{amsthm}
\usepackage{hyperref}
\usepackage{footnote}
\usepackage{nicefrac}       
\usepackage{xcolor}         
\usepackage{diagbox}
\usepackage{wrapfig,lipsum}
\usepackage{mathtools}
\usepackage{blindtext}
\usepackage{makecell}
\usepackage{longtable}
\usepackage{tabu}
\usepackage{soul} 

\usepackage{upgreek}
\usepackage[switch]{lineno}  

\begin{document}

\title{Improving Non-autoregressive Translation Quality with Pretrained Language Model, Embedding Distillation and Upsampling Strategy for CTC}

\author{Shen-sian Syu, Juncheng Xie, Hung-yi Lee %
        \thanks{National Taiwan University, Taiwan. They are in Taipei, TW.
        Email:\texttt{\{d07921013,f07942150,hungyilee\}@ntu.edu.tw }}

\thanks{This paper was published in the IEEE/ACM TRANSACTIONS ON AUDIO, SPEECH, AND LANGUAGE PROCESSING, VOL. 32, 2024. DOI: 10.1109/TASLP.2024.3451977.  The copyright has been transferred to IEEE.}
}

\maketitle

\begin{abstract}
Non-autoregressive approaches, especially those that generate output in a one-pass forward manner, have shown great potential in improving the inference speed of translation models. However, these approaches often suffer from a significant drop in translation quality compared to autoregressive models (AT).

To tackle this challenge, this paper introduces a series of innovative techniques to enhance the translation quality of non-autoregressive neural machine translation (NAT) models while still maintaining a substantial acceleration in inference speed. Specifically, we propose a method called CTCPMLM, which involves fine-tuning Pretrained Multilingual Language Models (PMLMs) with the Connectionist Temporal Classification (CTC) loss to effectively train NAT models. Additionally, we adopt the MASK insertion scheme instead of token duplication for up-sampling and present an embedding distillation method to further enhance the performance of NAT models.

In our experiments, CTCPMLM surpasses the performance of the baseline autoregressive model (Transformer \textit{base}) on various datasets, including WMT'14 DE$\leftrightarrow$EN, WMT'16 RO$\leftrightarrow$EN, and IWSLT'14 DE$\leftrightarrow$EN. Moreover, CTCPMLM represents the current state-of-the-art among NAT models. Notably, our model achieves superior results compared to the baseline autoregressive model on the IWSLT'14 En$\leftrightarrow$De and WMT'16 En$\leftrightarrow$Ro datasets, even without using distillation data during training. Particularly, on the IWSLT'14 DE$\rightarrow$EN dataset, our model achieves an impressive BLEU score of 39.93, surpassing AT models and establishing a new state-of-the-art. Additionally, our model exhibits a remarkable speed improvement of 16.35 times compared to the autoregressive model.
\end{abstract}

\IEEEpeerreviewmaketitle

\section{Introduction} \label{intro}
\IEEEPARstart{A}{utoregressive} (AT) machine translation\cite{mikolov2010recurrent,bahdanau2014neural,vaswani2017attention} models have long been the state-of-the-arts on various translation tasks, but one of the major drawbacks is not efficiently parallelizable on GPU or TPU during inference. To solve this problem, the prior work \cite{Gu0XLS18} proposed the non-autoregressive neural machine translation (NAT) models, which can generate all target tokens in parallel. While the speedup is significant, NAT suffers from degraded generation quality compared to AT models.

One of the main challenges is determining the appropriate target sequence length in advance. In the case, \cite{Gu0XLS18} employs Noisy Parallel Decoding (NPD), which generates multiple predictions of different lengths with fertility, and then a pretrained AT model is used to rescore them. \cite{stern19a,GuWZ19}  applied insertion method, which is partially autoregressive and uses the insertion operations for more flexible and dynamic length changes. \cite{shu2020latent} refines the latent variables instead of the tokens, allowing for dynamically adaptive prediction length. One popular solution is the CTC-based NAT models \cite{libovicky-helcl-2018-end, saharia-etal-2020-non,gu-kong-2021-fully,shao-etal-2022-one,shao2022nmla,wang-etal-2022-xlm,liang-etal-2023-dynamic}, which introduces connectionist temporal classification (CTC) \cite{GravesFGS06} to solve the target length problem. 

Another main challenge is the multi-modality problem. NAT removes the conditional dependence between target tokens, and predicting the independent issue leads to multi-modal outputs and induces the repetition of tokens~\cite{Gu0XLS18,zhou2021understanding}. A standard approach to overcome the multi-modality problem is to use sequence-level knowledge distillation \cite{kim-rush-2016-sequence}, by replacing the target side of the training set with the output from an AT model. Various advanced model architectures \cite{lee-etal-2018-deterministic,ghazvininejad-etal-2019-mask,xiao2023amomadaptivemaskingmasking} use iteration-based methods that can see the previous or partially predicted tokens to add dependence to solve the multi-modality problem.

There have been some pretrained encoder-decoder-based multilingual language models \cite{lample2019cross,zanon-boito-etal-2020-mass,liu-etal-2020-multilingual-denoising,xue-etal-2021-mt5,li-etal-2022-universal} that achieve superior translation performance. Most of them are dedicated to AT, while \cite{li-etal-2022-universal} intends to pretrain models that can be applied to both AT and NAT. Some \cite{rothe-etal-2020-leveraging, lample2019cross, Zhu2020Incorporating, 10.5555/3495724.3496634} otherwise incorporate encoder-based pretrained multilingual language models (PMLMs) into encoder-decoder-based transformer AT models for better translation performance.\cite{li-etal-2022-universal, Zhu2020Incorporating, 10.5555/3495724.3496634,su-etal-2021-non,wang-etal-2022-xlm,liang-etal-2023-dynamic} utilizes PMLMs to initialize NAT models. Among them, \cite{li-etal-2022-universal, Zhu2020Incorporating, 10.5555/3495724.3496634,wang-etal-2022-xlm,liang-etal-2023-dynamic} are iterative models with strong performance while \cite{su-etal-2021-non} is one-pass with weaker performance.


Based on the aforementioned findings, our goal is to integrate these techniques and propose a novel method to enhance the performance of NAT. In this paper, we introduce CTCPMLM, a one-pass NAT approach that combines the utilization of encoder-based PMLMs for model initialization and the integration of the CTC loss with sequence-level knowledge distillation using a strong AT teacher. Moreover, in the context of CTC, the input length must surpass the output length, necessitating upsampling. We propose an upsampling method involving the Fixed Ratio (FR) and Dynamic Ratio (DR). Additionally, we introduce an Insertion Method to handle the incorporation of additional tokens during upsampling.
Furthermore, we employ knowledge distillation to distill embeddings from the target language layer of the PMLM, enabling us to retain knowledge from the PMLM.

In summary, our contributions are: 
\begin{itemize}
    \item Unlike previous works \cite{rothe-etal-2020-leveraging, 10.5555/3495724.3496634, Zhu2020Incorporating,su-etal-2021-non,wang-etal-2022-xlm,liang-etal-2023-dynamic} that incorporates PMLMs with additional modules, we directly finetune PMLMs with CTC without introducing additional parameters, verifying it is an efficient and effective method in experiments.

    \item We propose a new upsampling scheme of inserting "MASK" tokens and an adaptive upsampling rate that further increases the performance of our NAT models. We also conduct a thorough search on different upsampling schemes to make the model achieve better accuracy.

    \item We propose to distill the knowledge from a frozen PMLM in the contextualized embedding level as a regularization method for our NAT model to retain the target language information learned from pretraining.
    
    \item Our model outperforms the baseline auto-regressive model (Transformer \textit{base}) on the WMT'14 DE$\leftrightarrow$EN, WMT'16 RO$\leftrightarrow$EN, and IWSLT'14 DE$\leftrightarrow$EN datasets, achieving a speed improvement of 16.35 times compared to the auto-regressive model. Additionally, the CTCPMLM represents the state-of-the-art among NAT models.

    \item Our model achieves a BLEU score of 39.93 on IWSLT'14 DE$\rightarrow$EN, which is a new state-of-the-art performance. It is worth noting that our performance surpasses that of baseline AT models even on IWSLT'14 En$\leftrightarrow$De and WMT'16 En$\leftrightarrow$Ro datasets, even when trained on the raw data (undistilled training set).
\end{itemize}

\section{Background} \label{background}
\subsection{Autoregressive Neural Machine Translation}
The current state-of-the-art Neural Machine Translation (NMT) models are autoregressive – generating sequences based on a left-to-right factorization. The output distributions are conditioned on the previously generated tokens \cite{VaswaniSPUJGKP17}. Let $x$ denote a source sequence and $y$ denote a target sequence, where $y_i \in \mathcal{V}$ and $\mathcal{V}$ is the target vocabulary. The joint probability is calculated as follows:
\begin{equation}
\begin{split}
    p_\theta(y\vert x)=\prod_{y_i\in y}p(y_i\vert y_{<i},x,\theta),
\end{split}
\end{equation}
where $y_{<i}$ denotes the previously generated tokens. Here, the probability of emitting each token $p(y_i\vert y_{<i},x,\theta)$ is parameterized by $\theta$ with a autoregressive model. This property of generated sequences based on a left-to-right factorization process is hard to be parallelized to make efficient use of computational resources and increases translation latency.

\subsection{Non-Autoregressive NMT}
The output distribution of non-autoregressive models is used conditionally independently among each token. The probability of the target sequence is modeled as follows:

\begin{equation}
\begin{split}
    p_\theta(y\vert x)=\prod_{y_i\in y}p(y_i\vert x,\theta)
\end{split}
\end{equation}

In contrast to the autoregressive model, the non-autoregressive model can emit tokens simultaneously and is easier to parallelize and reduce translation latency.
The main challenge of non-autoregressive NMT is the independence assumption that has a negative impact on translation quality. The significant limitations of non-autoregressive NMT currently suffer from (1) multimodality problem \cite{Gu0XLS18}, and (2) the inflexibility of prefixed output length \cite{Gu0XLS18,ghazvininejad-etal-2019-mask,https://doi.org/10.48550/arxiv.2204.09269}.

\subsection{Connectionist Temporal Classification (CTC)}
The vanilla non-autoregressive NMT \cite{Gu0XLS18} needs to predict the target length to construct the decoder input. To get better performance, \cite{Gu0XLS18} propose to generate multiple candidates with different lengths and re-ranking them to get the final translation that needs more computing power to produce multiple translations. This issue can use CTC \cite{GravesCTC06, Graves2013CTC} to solve. CTC models generate the alignment with repeated tokens and blank tokens. 
We can calculate the log-likelihood loss:
\begin{equation}
\begin{split}
\log p(y\vert x,\theta)=-\log\sum_{a\in \Gamma(y)}p(a\vert x,\theta),
\end{split}
\end{equation}
where $\Gamma(y)$ returns all possible alignments for a sequence $y$ with repeated tokens and blank tokens, given a sequence $x$ of length $|x|$ and $y$ of length $L_y$, where $|x| \ge |y|$.
 The alignment will be post-processed by a collapsing function $\Gamma(y)^{-1}$ to remove all blanks and collapses consecutive repeated tokens.
CTC assumes the source sequence is at least as long as the target sequence~\cite{libovicky-helcl-2018-end, saharia-etal-2020-non}. However, this principle is not innately implemented in the field of machine translation, which makes the application of CTC to machine translation nontrivial.
\subsection{Upsampling Method} 
\begin{figure}[htbp]
    \centering
    \includegraphics[width=0.5\textwidth]{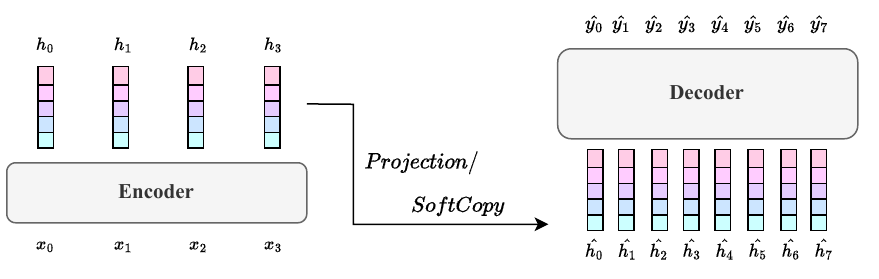}
    \caption{The upsampling method solves the problem that the length of the generated sentence needs to be longer than the source sentence. s=2}
    \label{fig:bk_upsampling_method}
\end{figure}

To solve the length of generate sentence need to longer the source sentence. One possible solution is to use a trainable projection layer to upsample the hidden state $h$ by an upsampling rate $s$\cite{libovicky-helcl-2018-end}. This involves applying a linear projection, which divides each state into $s$ vectors, resulting in a sequence length change from $|x|$ to $s|x|$. Another approach is utilizing $SoftCopy$~\cite{wei-etal-2019-imitation,gu-kong-2021-fully}, where attention weights $w_{ij}=\text{softmax}(-|j-i|/\tau)$ depend on the distance relationship between the source position $i$ and the upsampled position $j$ to calculate the upsampling vector $\hat{h}_j=\sum_{i=0}^{|x|}w_{ij}h_i$, with $\tau$ being a hyperparameter to adjust the degree of focus during copying. Figure \ref{fig:bk_upsampling_method} provides an illustrative diagram.

\section{Approach} \label{approach}
\begin{figure*}[ht]
    \centering
    \includegraphics[width=0.7\textwidth]{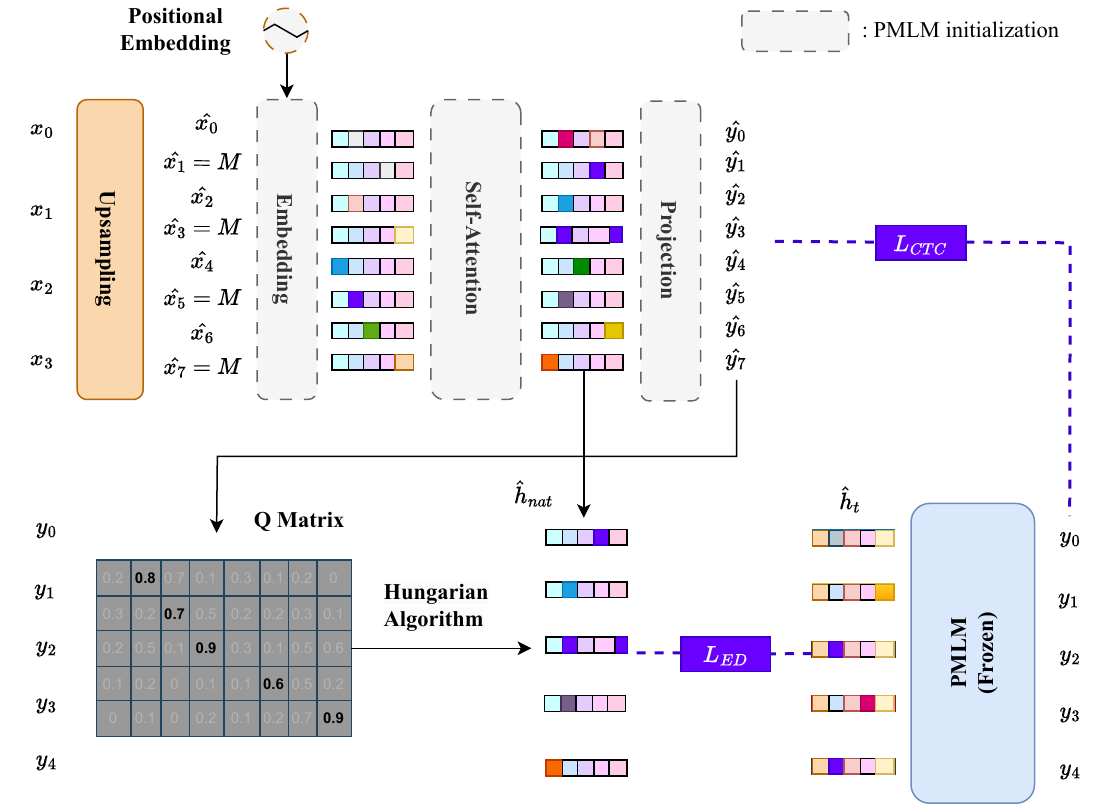}
    \caption{The overall framework of our CTCPMLM NAT model. The label $M$ means MASK token.}
    \label{fig:model-arch}
\end{figure*}

In this section, we provide a detailed explanation of the proposed model. First, we describe how to utilize pre-trained language models (PMLMs) as non-autoregressive translation (NAT) models. Second, we discuss the various CTC upsampling methods and propose a dynamic strategy to determine the upsampling ratio for each sentence.
Next, to maintain the knowledge embedded in the PMLM, we introduce a PMLM teacher to guide the NAT student model in aligning the PMLM embedding space, a process we call embedding distillation (ED).
The overall framework of training our CTCPMLM system is presented in Figure \ref{fig:model-arch}, where $x$, $\hat{x}$, $y$ and $\hat{y}$ denote the source sentence, the source sentence after upsampling, the target sentence and the model prediction respectively. The symbols $h_{nat}$, $\hat{h}_{nat}$, and $\hat{h}_t$ represent the last hidden state of the NAT model, the output obtained using the Hungarian algorithm, and the hidden state of the frozen PMLM, respectively.

\subsection{PMLM Initialization}
Many works \cite{li-etal-2022-universal,su-etal-2021-non, Zhu2020Incorporating, 10.5555/3495724.3496634,wang-etal-2022-xlm,liang-etal-2023-dynamic} have successfully used pre-trained language models (PLMs) to initialize NAT models for machine translation tasks. Inspired by these works, we directly fine-tuned the PLMs with CTC loss to leverage the learned knowledge during the pre-training stage. Our model is a transformer encoder-only PLM divided into three parts for training: word embedding layer, self-attention layer, and output projection layer, as shown in Figure \ref{fig:model-arch}.

\subsection{Upsampling Strategy for CTC} 
In this study, we utilize the connectionist temporal classification (CTC) loss~\cite{graves2006connectionist,chan2020imputer}, also referred to as latent alignment, for model training. One major benefit of the CTC loss function is that it does not require predicting the target sequence length. However, CTC necessitates that the input length be greater than the output length. In translation tasks, the source language is not always longer than the target language, making upsampling of the input necessary. To address this, we propose two upsampling strategies: the ratio upsampling method and the insertion method. These strategies are incorporated into the "Upsampling" function block in Figure~\ref{fig:model-arch}.

\paragraph{Ratio of Upsampling} 
Next, we introduce two types of upsampling ratios: Fixed Ratio (FR) and Dynamic Ratio (DR). 
\begin{itemize}
    \item Fixed Ratio (FR):
    The simplest method is to use the most likely choices for the upsampling ratio. Specifically, we multiply the length of each source sentence $|x|$ by a fixed factor $s$ to ensure that $s|x|$ is greater than the output length $L_y$. To avoid the position values exceeding the maximum position $pos_{\text{max}}$ (e.g. 512 for mBERT) of the PLM after upsampling, we clip the upsampling position values, which is $position = \min(i, pos_{\text{max}}) , i =[0,s|x|-1$].
    \item Dynamic Ratio (DR):
        Although a FR can be used for upsampling, it may lead to the loss of positional information beyond $pos_{\text{max}}$. A simple solution is to reduce the ratio $s$, but this could degrade output quality. This is because a higher upsampling ratio (e.g., $s=4$) offers a larger alignment space, resulting in improved accuracy \cite{gu-kong-2021-fully}. To prevent information loss in long sentences after upsampling, we propose the DR approach in addition to the FR method. With DR, we dynamically adjust the upsampling ratio based on the length of the source sentence $pos_{\text{max}}$, ensuring that long input sentences have lengths below $l_{pos}$ after upsampling. 
        The formula for calculating the $s_{DR}$ ratio is as follows:
        \begin{equation}
        s_{DR}= \left\{ 
            \begin{array}{lcl} 
                s & \text{for} & s|x| \leq pos_{\text{max}} \\
                \frac{pos_{\text{max}}}{|x|} & \text{for} & s|x| > pos_{\text{max}}
            \end{array}
        \right.
        \end{equation}
    
\end{itemize}



\paragraph{Insertion Method} Following the CTC setup of \cite{libovicky-helcl-2018-end, saharia-etal-2020-non}, we set upsampled source sentence $\hat{x}$ to be $s$ times the length of the original source sentence $x$.
To reduce the disparity between the token embeddings of our PMLM and its pretraining, unlike prior approaches \cite{libovicky-helcl-2018-end,wei-etal-2019-imitation,gu-kong-2021-fully}, we do not use the projection layer or SoftCopy method. Instead, our upsampling can be done by duplicating source tokens (IT, "Inserting Tokens" in the following sections) or inserting the [MASK] tokens (IM, "Inserting Masks" in the following sections), which is the special “MASK” token used during BERT pretraining. For instance, given source sequence $x= [A, B, C, D]$ and $s=2$, it will be $\hat{x}=[A,A,B,B,C,C,D,D]$ by IT, and $\hat{x}=[A,M,B,M,C,M,D,M]$ by IM. 

When dealing with floating-point numbers as the upsampling ratio $s$, the upsampled sequence $\hat{x}$ might have a non-integer length. To address this issue, we use the following formula:

\begin{equation}
\hat{x}_j = \begin{array}{lcl} 
x_i & \text{for } i \text{ where } \arg\min\limits_{i}\left| j - (i+1) \times s + \frac{s}{2} \right|
\end{array}
\end{equation}

This formula essentially identifies the nearest "insertion point" around position $j$, determined by the upsampling rate. Here, $i$ represents the position of the source token, ranging from $0$ to $|x|$, and $j$ represents the position of the upsampled token, ranging from $0$ to $\lfloor s|x| \rfloor$. The approach for the IM Method is similar, with the difference that originally duplicated tokens are replaced with the [MASK] tokens. 

Formally, given the upsampled source sequence $\hat{x}$ and target sequence $y$, the proposed NAT objective is defined as:

\begin{equation}
\begin{split}
L_{CTC}=-\log\sum_{a\in \Gamma(y)}p(a\vert \hat{x}, \theta) 
\end{split}
\end{equation}

\subsection{Embedding Distillation (ED)}
Using PMLM for model initialization enables aligning the target-side space by obtaining contextualized representations from the target language. The goal is to improve the target contextualized representations for NAT models, approximating those of the teacher model. The teacher model, represented as "PMLM(Frozen)" in Figure \ref{fig:model-arch}, is the same PMLM as the NAT model, with frozen parameters during training. Our approach involves maximizing the cosine similarity between the representations of the NAT model and the frozen PMLM.

\begin{equation}
\begin{split}
sim(h_{nat},h_t^l)=\frac{(h_{nat} {h_t^l}^T)}{(\|h_{nat}\| \|h_t^l \|)}
\end{split}
\end{equation}

where $h_{nat}$ is the representation from the last layer of the NAT model with the source sequence as input, and $h_t^l$ is the representation from the $l$-th layer of the teacher model with the target sequence as input.
Due to the different lengths of $h_{nat}$ and $h_{t}^l$, and the possibility that the predicted sequences $\hat{y}$ and target sequences $y$ may not be monotonic, it is necessary to determine the alignment between $\hat{y}$ and $y$ in order to compute the similarity of representations of the NAT model and the frozen PMLM based on this alignment. Here, we first compute the probability matrix $Q_{ij}=\log p(y_i\vert \hat{h}_{nat_j}), \forall i,j ,i=0\cdots\vert y \vert, j=0\cdots s|x| $, $Q \in \mathbb{R}^{\vert y \vert \times s \vert x \vert}$. The matrix $Q$ represents the probability of each target token at each position of the predicted token.  Then, we use the Hungarian algorithm\cite{https://doi.org/10.1002/nav.3800020109,pmlr-v139-du21c} to calculate the alignment based on the probability matrix $Q$, and find the optimal matching pairs ($\hat{h}_{nat_j}, h^l_{t_i}$) based on the alignment between $\hat{y}$ and $y$. Subsequently, we calculate token distilled loss which is defined as: 
\begin{equation}
\begin{split}
      L_{ED}=\frac{\sum_{i=0}^{|y|}(1-sim(\hat{h}_{nat_j} , h^l_{t_i}))}{|y|}
\end{split} 
\end{equation}

\subsection{Final Objective}In summary, our training objective is a combination of the proposed objectives, and we jointly train the model with them at each training step:
\begin{equation}
\label{eqn:lctc}
\begin{split}
      L= L_{CTC}+ \lambda L_{ED}
\end{split} 
\end{equation}
Here, $\lambda$ is a hyperparameter, which is set to 0 or 1 in our experiments. Further training details are described in the Experiment section.

\section{Experiment} \label{experiment}
\subsection{Experiment Setup}
\paragraph{Datasets}  We evaluate the proposed NAT on three widely used public machine translation corpora: IWSLT'14 En$\leftrightarrow$De, WMT'16 En$\leftrightarrow$Ro, and WMT'14 En$\leftrightarrow$De. We use the low-resourced dataset - IWSLT'14 for hyperparameter search and ablation studies. The validation and test set for WMT'16 En$\leftrightarrow$Ro are \verb|newsdev-2016| and \verb|newstest-2016| respectively. We follow \cite{xu-etal-2021-bert} and convert WMT'14 En$\leftrightarrow$De to lowercase.

For WMT'14 En$\leftrightarrow$De, the test set is \verb|newstest2014| while we follow \cite{xu-etal-2021-bert} and combine \verb|newstest2012| and \verb|newstest2013| for validation. The statistics of all 3 corpora are summarized in TABLE~\ref{tab:data-static}. We adopt the same preprocessing steps as \cite{gu-kong-2021-fully} for WMT'16 En$\leftrightarrow$Ro and WMT'14 En$\leftrightarrow$De. All the data are tokenized by the Huggingface tokenizer\footnote{\url{https://huggingface.co/docs/transformers/main_classes/tokenizer}} associated with the PMLM.

\begin{table}[htbp]
    \centering
    \scriptsize
    \caption{The statistics of all 3 corpora }
    \begin{tabular}{cccc}
        \toprule
        Dataset Statistics & Train & Valid & Test \\
        \midrule
        IWSLT'14\cite{cettolo-etal-2014-report}\ EN$\leftrightarrow$DE & 160,239 & 7,283 & 6,750 \\
        WMT'16\cite{bojar-etal-2016-findings} EN$\leftrightarrow$RO & 608,319 & 1,999 & 1,999 \\
        WMT'14\cite{bojar-etal-2014-findings} EN$\leftrightarrow$DE & 3,961,179 & 6,003 & 3,003 \\
    \bottomrule
    \end{tabular}
    \label{tab:data-static}
\end{table}
%
%

\paragraph{Knowledge Distillation (KD) Teachers}  
From the paper \cite{kim-rush-2016-sequence}, sequence-level knowledge distillation can assist in generating less complex and less noisy training data for NAT student models. To assess the impact of different teachers on NAT performance, we classify the distillation teachers into two groups: the base model and the strong model. The base model employs the standard base architectures \cite{VaswaniSPUJGKP17}, denoted as transformer \textit{base} in subsequent sections. The strong model consists of powerful AT models that achieve higher BLEU scores on the corresponding datasets. Now, we will describe the specific strong teachers used for different datasets:
\begin{itemize}
    \item For IWSLT'14 En$\leftrightarrow$De and WMT'14 En$\leftrightarrow$De, we use the state-of-the-art model proposed by \cite{xu-etal-2021-bert}, which is an AT transformer with a frozen BiBERT PMLM as its contextualized input embeddings.
    \item For WMT'16 En$\leftrightarrow$Ro, we utilize the strong model from \cite{bhosale-etal-2020-language}, both of which are trained with back-translation data\footnote{\url{https://data.statmt.org/rsennrich/WMT'16_backtranslations/}}.
\end{itemize}

\paragraph{Backbone PMLMs} We utilize encoder-based PMLMs along with their output projection layer as our NAT models. For experiments on IWSLT'14 En$\leftrightarrow$De and WMT'14 En$\leftrightarrow$De dataset, the NAT model is BiBERT (\verb|jhu-clsp/bibert-ende|) pretrained by \cite{xu-etal-2021-bert} on monolingual data of English and German. For WMT'16 Ro$\leftrightarrow$En, we use mBERT (\verb|bert-base-multilingual-uncased|) pretrained by \cite{devlin-etal-2019-bert}. 

\paragraph{Training} We train IWSLT'14 En$\leftrightarrow$De baseline AT teacher models following \cite{ott2019fairseq}. For baseline AT teachers on WMT'16 Ro$\leftrightarrow$En and WMT'14 En$\leftrightarrow$De, we use the distillation data by \cite{gu-kong-2021-fully}\footnote{\url{https://github.com/shawnkx/Fully-NAT}}. We measure the validation BLEU scores for every epoch and average the best 5 checkpoints to obtain the final model. We implement our models based on fairseq \cite{ott2019fairseq} (MIT license) and Huggingface \cite{wolf-etal-2020-transformers}\footnote{\url{https://github.com/huggingface/transformers}} (Apache-2.0 license). For our NAT models, the dropout rates are all 0.1 and learning rates are all $10^{-4}$. For the IWSLT'14 En$\leftrightarrow$De task, the NAT models are trained with a batch size of 12.8k tokens for 50K updates. For the WMT'16 Ro$\leftrightarrow$En task, the NAT models are trained with a batch size of 65K tokens for 30K updates. Finally, for the WMT'14 En$\leftrightarrow$De task, the NAT models are trained with a batch size of 65k tokens for 100K updates. In equation \ref{eqn:lctc}, the parameter $\lambda$ is initially set to 0 for the first 30k steps on the IWSLT'14 En$\leftrightarrow$De task, and then set to 1 for the remaining steps. Similarly, for the WMT'14 En$\leftrightarrow$De task, $\lambda$ is set to 0 for the first 75k steps and then to 1 for the remaining steps. For the WMT'16 Ro$\leftrightarrow$En task, $\lambda$ is set to 0 for the first 20k steps and then to 1 for the remaining steps. We analyze the training cost on three datasets, and the results are presented in TABLE~\ref{tab:train-cost}.
\begin{table}[ht]
\centering
\scriptsize
\caption{The statistics of CTCPMLM training. ’Model’ means the PMLM of our NAT models. ’Step’ means the number of training steps. ’Batch’ means number of tokens in a batch. ’Time’ means the training time measured on 4 GeForce RTX 3090 GPUs.
}
\begin{tabular}{lcccc}
\toprule
Dataset & Model & Step & Batch & Time \\
\midrule
IWSLT'14 En$\leftrightarrow$De  & BiBERT & 50k & 12288 & 5.9hrs \\
WMT'14 En$\leftrightarrow$De & BiBERT & 100k & 65536 & 43.5hrs \\
WMT'16 Ro$\leftrightarrow$En & mBERT  & 30k & 65536 & 15.40hrs \\
\bottomrule
\end{tabular}

\label{tab:train-cost}
\end{table}

\paragraph{Evaluation} Following \cite{papineni-etal-2002-bleu}, we evaluate the performance of our models with the tokenized BLEU score. The translation latency measurement is done on a single Nvidia 3090 GPU with one sentence at a time.

\paragraph{Decoding} Following \cite{gu-kong-2021-fully,shao2022nmla}, CTC-based NAT models can also be combined with 4-gram language models \cite{heafield-2011-kenlm}\footnote{\url{https://github.com/kpu/kenlm}} (LGPL license) in CTC beam search decoding for the optimal translation $y$ that maximizes:
\begin{equation}
\label{eqn:ctcbeam}
\begin{split}
    \log p_\theta(y\vert x) + \alpha \log p_{LM}(y) + \beta \log p(\vert y \vert)
\end{split}
\end{equation}
where $\alpha$ and $\beta$ are hyperparameters for the weights of language model score and length bonus. Since there is no neural network computation in its algorithm, it can be efficiently implemented in C++\footnote{\url{https://github.com/parlance/ctcdecode}} (MIT license). In the following experiment, we use fixed $(\alpha , \beta) = (0.3, 0.9)$ and beam size 20.

\subsection{Hyperparameter Search}
We do the hyperparameter search on the validation set of IWSLT'14 De$\rightarrow$En for the following 5 settings: (1) Frozen/Trainable Word Embedding layer, (2) Frozen/Trainable output projection layer, (3) Inserting duplicated token (IT) or inserting ”MASK” (IM) for upsampling, (4) Upsampling Ratio $s$, (5) Dynamic (DR) or Fixed (FR) upsampling ratio.
From the results of the following experiments, we find the NAT with a frozen word embedding, a trainable output projection layer, inserting masks, and using a dynamic upsampling ratio with $s=4$ to obtain the best performance on the validation set.

\begin{figure}[htbp]
    \centering
    \includegraphics[width=0.45\textwidth]{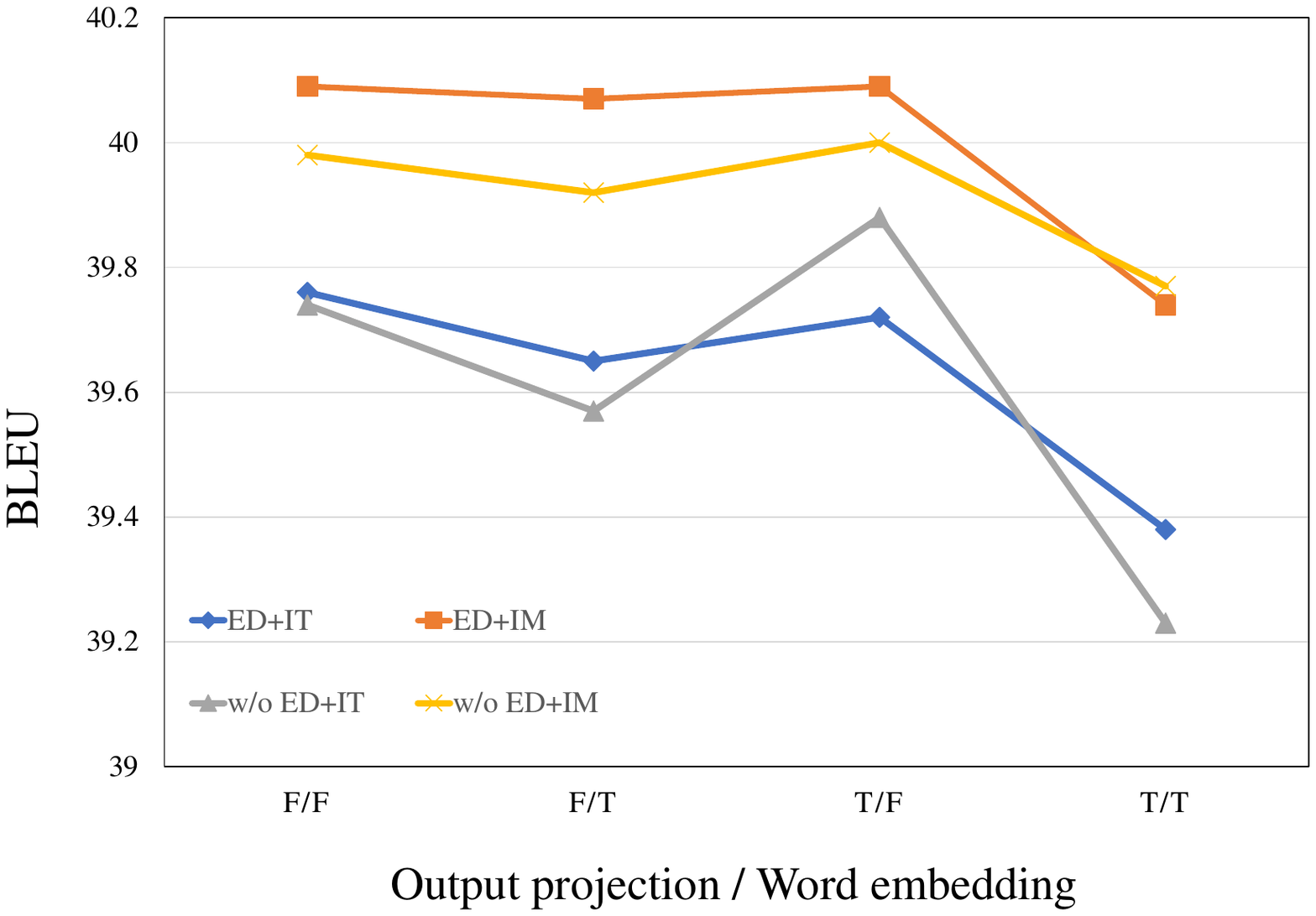}
    \caption{BLEU v.s. Frozen(F) or Trainable(T) parameters on the validation set of IWSLT'14 DE$\rightarrow$EN. We evaluate the settings of frozen or trainable output projection layer and word embedding layer on 4 different conditions with upsampling ratio $s=2$ and FR.}
    \label{fig:LM Head and SWE Freeze_Trainable}
\end{figure}

\paragraph{Word embedding and output projection layers} In \cite{xu-etal-2021-bert}, the encoder of an AT model takes the contextualized embeddings from various PMLMs as input. Inspired by this idea, we try to freeze the parameters of input word embedding of our NAT models during training for the PMLM NAT model to retain the information of each token in its embedding space learned from pretraining. In addition, we also try to freeze the output projection layer of the NAT model to keep its representations in the target space close to the original representations of the pretrained PMLM. We combine two conditions of whether to freeze the input embeddings/output projection layer with other two conditions. One is the choice between IT and IM. Another is whether to apply ED. We try 16 different settings with upsampling ratio $s=2$ and FR. Figure \ref{fig:LM Head and SWE Freeze_Trainable} shows the BLEU scores on different frozen parameters and insertion methods (IT or IM).
Based on the experimental findings, it was observed that the utilization of a frozen word embeddings layer and a trainable output projection layer led to improved performance. 
Consequently, this configuration will be adopted as the standard approach for both the main results and the ablation study.

\begin{figure}[ht]
    \centering
    \includegraphics[width=0.45\textwidth]{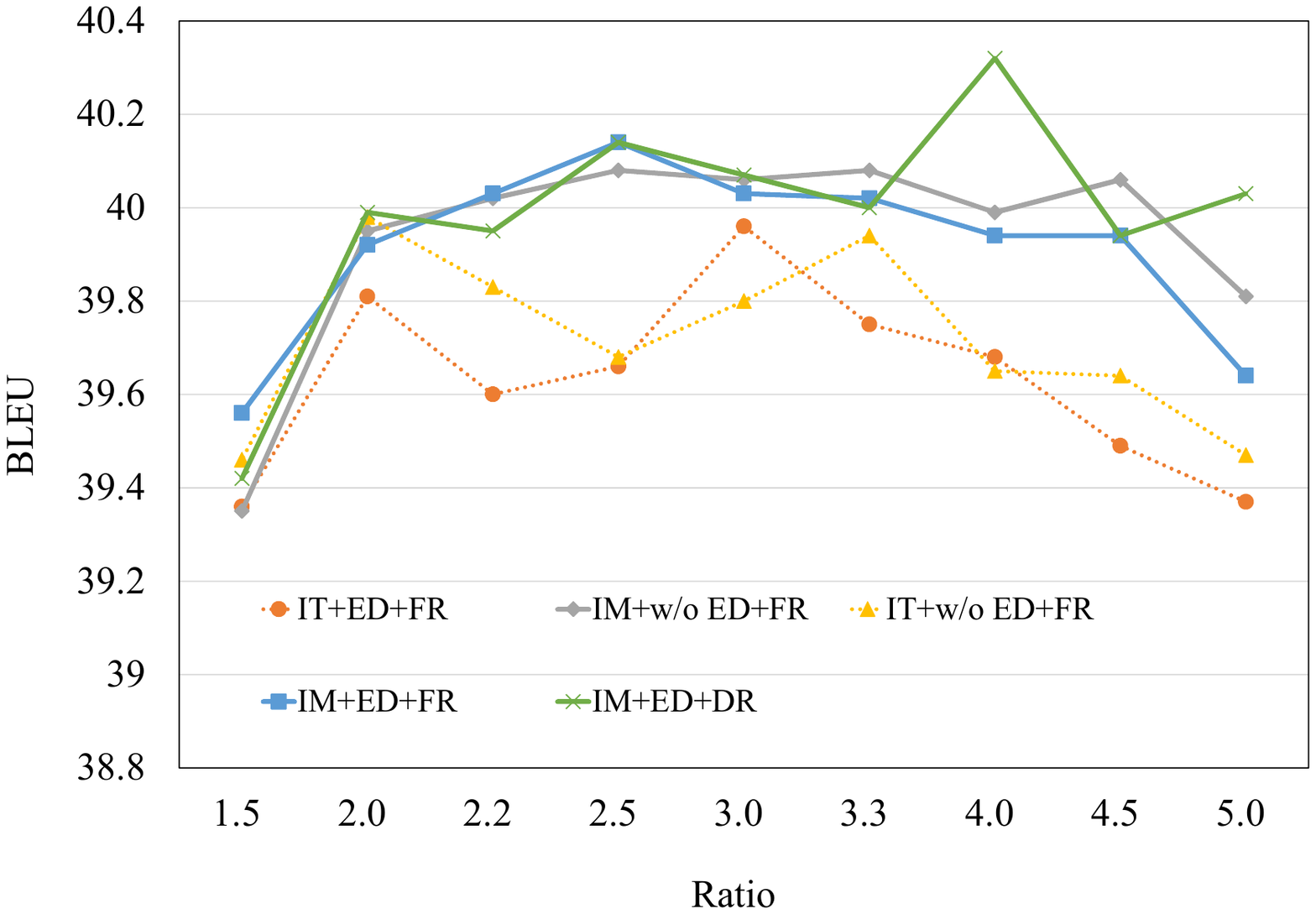}
    \caption{BLEU v.s. Upsampling Ratio: Parameter search on the validation set of IWSLT'14 DE→EN, investigating the evaluation of different upsampling strategies for CTC with or without ED, and utilizing BiBERT initialization.}
    \label{fig:diff_Rate_search}
\end{figure}

\paragraph{Upsampling Scheme and Upsampling Ratio}
In our experiment, we examine the impact of different upsampling rates and schemes (IT or IM), both with and without ED, using the BiBERT PLM for model initialization. The results, as shown in Figure \ref{fig:diff_Rate_search}, reveal several key observations:

\begin{itemize}
  \item Consistently, the use of the Inserting Masks (IM) method demonstrated superior performance compared to the Inserting Tokens (IT) method, this result is also evident in Figure \ref{fig:LM Head and SWE Freeze_Trainable}. Furthermore, when using IT, the BLEU score showed more significant variation across different upsampling ratios.
  \item Additionally, DR is particularly beneficial at higher upsampling rates, improving the performance of the model.
  \item Importantly, we observe that extremely low or high upsampling rates do not yield favorable results.
\end{itemize}

Considering these findings and the model parameters, we will adopt the approach of combining DR with IM and incorporating ED, while setting the upsampling rate to $s=4.0$ in the model initialized with BiBERT PMLM.

\begin{figure}[ht]
    \centering
    \includegraphics[width=0.45\textwidth]{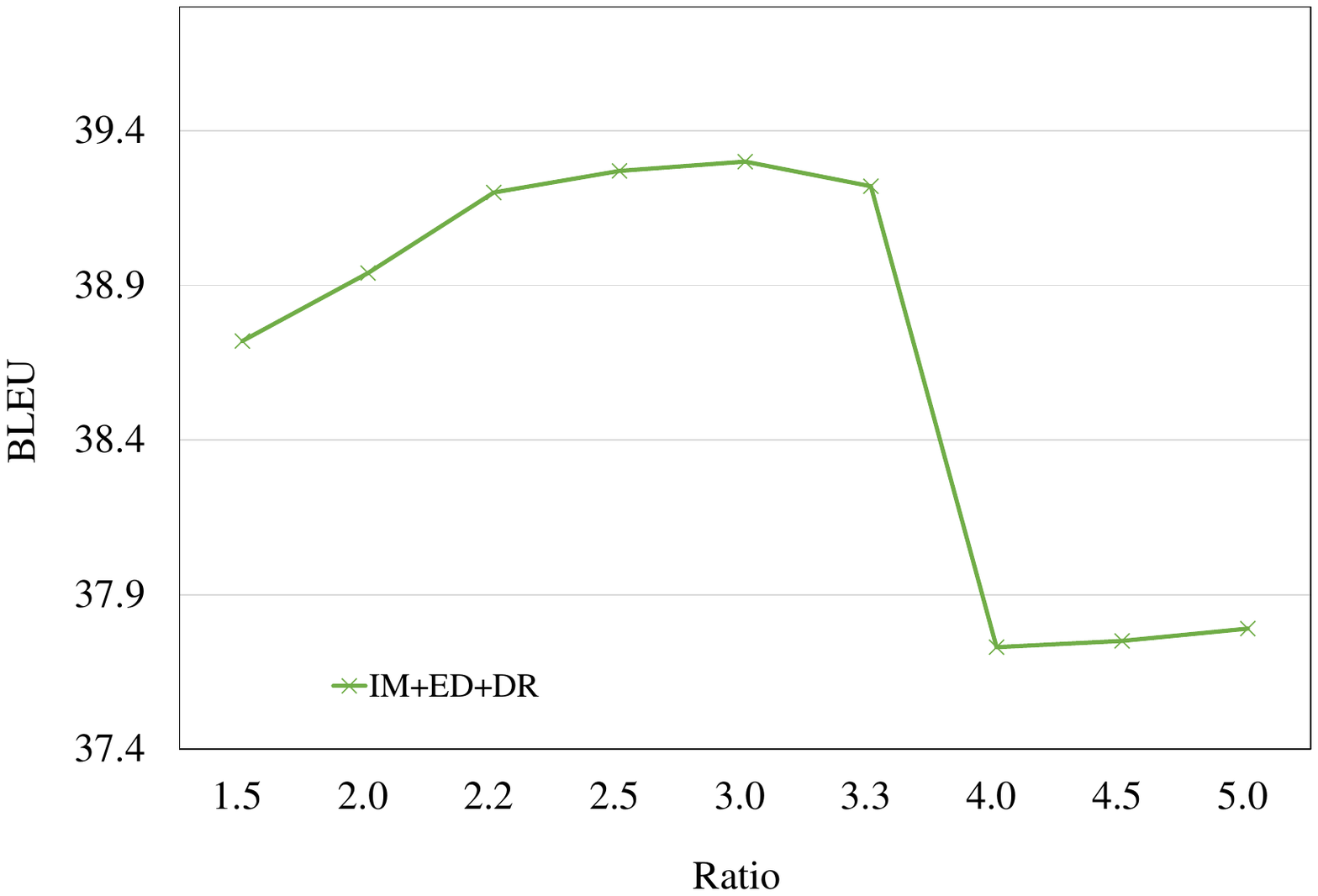}
    \caption{BLEU v.s. Upsampling Ratio: Parameter search on the validation set of IWSLT'14 DE→EN, investigating the impact of Upsampling Ratio on BLEU scores when evaluating the combined IM, ED, and DR strategies for mBERT initialization.}
    \label{fig:diff_rate_mBERT}
\end{figure}

The same hyperparameter search method is applied to determine the optimal upsampling ratio when initializing the model with mBERT PMLM. Figure \ref{fig:diff_rate_mBERT} also presents the validation set results on IWSLT'14 De$\rightarrow$En. Within the upsampling ratio range of 1.5 to 3.0, a gradual increase in the BLEU score is observed, eventually reaching a plateau. This suggests that increasing the upsampling ratio within this range can enhance translation quality. However, surpassing this range leads to a noticeable decrease in the BLEU score as the upsampling ratio continues to increase. 
Thus, surpassing a certain threshold does not yield improved results through higher upsampling. 
For the model initialized with mBERT, an upsampling ratio of $s=3.0$ was selected.

\subsection{Main Result}
Next, we will discuss each of these three datasets separately and analyze their characteristics and performance. For TABLE~\ref{tab:main-iwslt} and TABLE~\ref{tab:main-wmt}, we utilized the strong teacher models for our distillation data. 
\paragraph{IWSLT'14 De$\leftrightarrow$En}
Following a hyperparameter search on the validation set of IWSLT'14 De$\rightarrow$En, we apply a selection of the highest-performing settings to our NAT model and evaluated their performance on the test set. These results are presented in TABLE~\ref{tab:main-iwslt}. 
\begin{itemize}
    \item 
    The results in (a) and (b) clearly demonstrate the beneficial impact of Knowledge Distillation (KD) on enhancing the performance of the NAT model. Additionally, the comparison between (b) and (c) reveals a significant boost in the BLEU score, signifying a considerable improvement in the NAT model's performance compared to the randomly initialized model shown in row (b).
    Moreover, a comparison between rows (d) and (e) underscores the advantage of using mask tokens (IM) over duplicate tokens (IT) for the tokens newly inserted by upsampling. The results illustrate that using IM leads to a noticeable improvement in performance compared to using IT. Row (f) not only represents the optimal setting derived from our hyperparameter search but also showcases the best performance on the test set, thereby underlining the effectiveness of our model under this configuration.
    \item The introduction of CTC beam search in row (g) does not lead to performance improvement. In the En$\rightarrow$De translation direction, the BLEU score remains the same with no significant change compared to row (f). However, we observe a substantial performance drop in the De$\rightarrow$En direction.    
    \item Our model surpasses other one-pass or iterative NAT models in translation quality, regardless of whether a pre-trained model is used. Furthermore, our model also outperforms AT models, including those that are pre-trained. Notably, it even exceeds the performance of the teacher model. CTCPMLM sets a new state-of-the-art on this dataset, marking a significant advancement for NAT.
    \item In terms of speed, our model achieves a remarkable speedup of 16.35$\times$ compared to the baseline autoregressive model. Even when using beam search, our model still demonstrates a significant speed improvement of 7.16$\times$.
   
\end{itemize}

\begin{table*}[htbp]
\centering
\scriptsize
\caption{Performance of BLEU score on IWSLT'14 En$\leftrightarrow$De. "-" denotes that the results are not reported in the previous work. \textbf{Iter.} denotes the number of iterations at inference time. $^\Diamond$ denotes results from our run. $^\dag$ denotes trained without sequence-level knowledge distillation. Baseline NAT models use distillation data from base transformer teachers. "no-KD" means without sequence-level KD. "init" means the model is randomly initialized. The relative speedup of our models is measured on the test set of IWSLT'14 En$\rightarrow$De.}
\color{black} 
\begin{tabular}{llcrcc}
\toprule
 \multicolumn{2}{l}{\multirow{2}{*}{\textbf{Models}}} & 
 \multirow{2}{*}{\textbf{Iter.}} &
 \multirow{2}{*}{\textbf{Speed}}&
 \multicolumn{2}{c}{\textbf{IWSLT'14}}  \\
 \multicolumn{2}{c}{} & & & \textbf{EN-DE} & \textbf{DE-EN} \\
\midrule
\midrule
\textbf{Without Pre-trained} \\
\cmidrule[0.6pt](lr){1-6}
\multirow{1}{*}{AT}
& Transformer \textit{base} & N & 1.0$\times$ &  28.47 & 34.85 \\
\cmidrule[0.6pt](lr){1-6}
\multirow{2}{*}{Iterative NAT}
& JM-NAT~\cite{guo-etal-2020-jointly} & 10 & 5.7 $\times$ & - & 32.59 \\
& CMLMC~\cite{huang2021improving} & 10 & - & 28.51 & \textbf{34.78} \\
\cmidrule[0.6pt](lr){1-6}
\multirow{3}{*}{One-pass NAT}
& Glat~\cite{qian-etal-2021-glancing} & 1 & 15.3$\times$  & - & 29.61\\
& LAVA NAT~\cite{Li2020LAVANA} & 1 & 20.2$\times$  & - & 31.69\\
& DDRS + lm\&beam20~\cite{shao-etal-2022-one} & 1 & 5$\times$  & - & \textbf{34.74}\\

\midrule
\midrule
\textbf{With Pre-trained} \\
\cmidrule[0.6pt](lr){1-6}
\multirow{2}{*}{AT}
& BiBERT (teacher) \cite{xu-etal-2021-bert} & N & - & 30.36$^\Diamond$ & \textbf{38.22$^\Diamond$} \\
& BERT-Fuse~\cite{zhu2020incorporatingbertneuralmachine} & N & - & 30.45 & 36.11 \\

\cmidrule[0.6pt](lr){1-6}
\multirow{3}{*}{Iterative NAT}
& CeMAT~\cite{li-etal-2022-universal} & 10 & - & 26.70$^\dag$ & 33.70$^\dag$ \\
& AB-Net~\cite{10.5555/3495724.3496634} & - & 2.4$\times$ & - & 36.49$^\dag$ \\
& DEER~\cite{liang-etal-2023-dynamic} & 4 & 3.3$\times$ &   - & \textbf{37.46} \\
\cmidrule[0.6pt](lr){1-6}
\multirow{2}{*}{One-pass NAT}
& Bert+CRF-NAT~\cite{su-etal-2021-non} & 1 & 11.31$\times$  & - & 30.45\\
& DEER~\cite{liang-etal-2023-dynamic} & 1 & 12.0$\times$ &   - & \textbf{34.89} \\
\cmidrule[0.6pt](lr){1-6}

\multirow{7}{*}{CTCPMLM}

&  (a) no-KD + init + FR(s=3) + IT & 1 & 16.56$\times$ & 20.87$^\dag$ & 27.21$^\dag$  \\
&  (b)  init + FR(s=3) + IT & 1 & 16.56$\times$ & 26.66 & 33.30 \\
&  (c)  PLM + FR(s=3) + IT & 1& 16.56$\times$ & 31.60 & 39.52  \\
&  (d)  PLM + DR(s=4) + IT & 1 & 16.35$\times$& 31.82 & 39.45  \\
&  (e)  PLM + DR(s=4) + IM & 1 & 16.35$\times$& 31.89 & 39.68  \\
&  (f)  PLM + DR(s=4) + IM + ED & 1 & 16.35$\times$& \bf{31.92} & \bf{39.93}  \\
&  (g)  PLM + DR(s=4) + IM + ED + beam search & 1 & 7.16$\times$& 31.94 & 36.57 \\

\bottomrule
\end{tabular}
\label{tab:main-iwslt}
\end{table*}

\begin{table*}[htbp]
\centering
\scriptsize
\caption{
Performance of BLEU score on WMT'14 En$\leftrightarrow$De and WMT'16 En$\leftrightarrow$Ro. This notation "s=4/3" indicates $s=4$ in WMT'14 dataset, and $s=3$ in WMT'16 dataset respectively. "-" denotes that the results are not reported. 
The speedup and relative speedup are measured on WMT'14 De-En test set. 
\textbf{Iter.} denotes the number of iterations at inference time, \textbf{Adv.} means adaptive. $^\Diamond$ denotes results from our run. $^*$ denotes models trained with distillation from a \textit{transformer big} (8-1024-4096) model . $^\dag$ denotes trained without sequence-level knowledge distillation. Other NAT models use distillation data from base transformer teachers. $^\ddag$ denotes the strong knowledge distillation teacher for our models on WMT'16 En$\leftrightarrow$Ro, and their scores are tokenized BLEU calculated after their predictions are converted to lowercase. 
}
\color{black}
\begin{tabular}{llcrcccc}
\toprule
 \multicolumn{2}{l}{\multirow{2}{*}{\textbf{Models}}} & 
 \multirow{2}{*}{\textbf{Iter.}} &
 \multirow{2}{*}{\textbf{Speed}}&
 \multicolumn{2}{c}{\textbf{WMT'14}} & \multicolumn{2}{c}{\textbf{WMT'16}} \\
 \multicolumn{2}{c}{} & & & \textbf{EN-DE} & \textbf{DE-EN} & \textbf{EN-RO} & \textbf{RO-EN} \\
\midrule
\midrule
\textbf{Without Pre-trained} \\
\cmidrule[0.6pt](lr){1-8}
\multirow{4}{*}{AT}
& Transformer \textit{base} & N & 1.0$\times$ &  27.48 & 31.39 & 34.26 & 33.83 \\
& Transformer \textit{big}  & N & -           & \textbf{28.60}  & - &  & - \\
& FNC (\textit{beam=5}) \cite{bhosale-etal-2020-language} & N & - & - & - & - & \textbf{40.76$^\ddag$} \\
& Transformer \textit{big} + Backtranslation \cite{bhosale-etal-2020-language} & N & - & - & - & \textbf{41.61$^\ddag$} & - \\
\cmidrule[0.6pt](lr){1-8}
\multirow{5}{*}{Iterative NAT}
& CeMAT~\cite{li-etal-2022-universal} & 10 & - & 27.20 & 29.90 & 33.30$^\dag$ & 33.00$^\dag$ \\
& Disco~\cite{DBLP:journals/corr/abs-2001-05136} & Adaptive & 3.5$\times$ & 27.34$^*$ & 31.31$^*$ & 33.22 & 33.25 \\
& Rewrite-NAT~\cite{geng-etal-2021-learning} & 2.3 & 3.9$\times$ & 27.83$^*$ & 31.52$^*$ & 33.63 & 34.09 \\
& Imputer~\cite{saharia-etal-2020-non} & 8& 3.9$\times$ & \textbf{28.20$^*$} & \textbf{31.80$^*$} & \bf{34.40} & \bf{34.10} \\
& CMLM~\cite{ghazvininejad-etal-2019-mask} & 10 & 1.7$\times$ & 27.03$^*$ & 30.53$^*$ & 33.08 & 33.31 \\
& DA-Transformer ~\cite{huang2022directed} & 1 & 7.0$\times$ & 27.91$^*$ & 31.95$^*$  &  -& -\\
\cmidrule[0.6pt](lr){1-8}
\multirow{12}{*}{One-pass NAT}
& CTC~\cite{libovicky-helcl-2018-end} & 1 & - & 16.56 & 18.64 & 19.54 & 24.67 \\
& Vanilla-NAT~\cite{Gu0XLS18} &  1 & 15.6$\times$ & 17.69 & 21.47 & 27.29 & 29.06 \\

& NAT-REG~\cite{Wang_Tian_He_Qin_Zhai_Liu_2019} & 1 & 27.6$\times$ & 20.65 & 24.77 & - & - \\

& Glat~\cite{qian-etal-2021-glancing} & 1 & 15.3$\times$ & 25.21 & 29.84 & 31.19 & 32.04 \\

& NART-DCRF~\cite{NEURIPS2019_74563ba2} & 1 & 10.4$\times$ & 23.44 & 27.22 & - & - \\
& AligNART~\cite{song-etal-2021-alignart} & 1 & 13.2$\times$ & 26.40 &  30.40 & 32.50 &  33.10 \\
& OAXE-NAT~\cite{pmlr-v139-du21c} & 1 & 15.3$\times$ & 26.10$^*$ &  30.20$^*$ & 32.40 & 33.30 \\

& Imputer~\cite{saharia-etal-2020-non} & 1 & 18.6$\times$  & 25.80$^*$ & 28.40$^*$ & 32.30 & 31.70 
\\
& Fully-NAT(VAE)~\cite{gu-kong-2021-fully} & 1 & 16.8$\times$& 27.49 & 31.10 & 33.79 & 34.87\\
& Fully-NAT(GLAT)~\cite{gu-kong-2021-fully} & 1 & 16.8$\times$& 27.20 & 31.39 & 33.71 & 34.16\\
& DDRS+lm\&beam20~\cite{shao-etal-2022-one} & 1 & 5$\times$ & 28.33 & 32.43 & 35.42 & 35.81 \\
& NMLA+DDRS+lm\&beam20~\cite{shao2022nmla} & 1 & 5$\times$ & \bf{28.63} & \bf{32.65} & \bf{35.51} & \bf{35.85} \\
\midrule
\midrule
\textbf{With Pre-trained} \\
\cmidrule[0.6pt](lr){1-8}
\multirow{2}{*}{AT}
& BiBERT (teacher) \cite{xu-etal-2021-bert} & N & - & \textbf{30.98}$^\Diamond$ & \textbf{36.30$^\Diamond$} & - & - \\
& BERT-Fuse~\cite{zhu2020incorporatingbertneuralmachine} & N & - & 30.75& - & - & - \\
\cmidrule[0.6pt](lr){1-8}
\multirow{4}{*}{Iterative NAT}
& CeMAT~\cite{li-etal-2022-universal} & 10 & - & 27.20 & 29.90 & 33.30$^\dag$ & 33.00$^\dag$ \\
& AB-Net~\cite{10.5555/3495724.3496634} & - & 2.4$\times$ & 28.69$^*$ & \bf{33.57}$^*$ & - & \textbf{35.63} \\
& XLM-D(M=6)~\cite{wang-etal-2022-xlm} & 8 & 4.1$\times$ & \textbf{29.59} & 33.28 & \textbf{35.64} & 35.48 \\
& DEER\cite{liang-etal-2023-dynamic} & 4 & 3.3$\times$ & 28.56 & - & 35.53 & - \\
\cmidrule[0.6pt](lr){1-8}
\multirow{2}{*}{One-pass NAT}
& XLM-D with LT only~\cite{wang-etal-2022-xlm} & 1 & 19.9$\times$ & \textbf{27.46} & \textbf{30.68} & \textbf{34.70} & \textbf{34.29} \\
& DEER\cite{liang-etal-2023-dynamic} & 1 & 12.0$\times$ & 26.19 & - & 33.95 & - \\
\cmidrule[0.6pt](lr){1-8}
\multirow{7}{*}{CTCPMLM}
&  (h) no-KD + init + FR(s=3/3) + IT & 1 & 22.9$\times$ & 18.73$^\dag$ & 20.22$^\dag$ & 24.55$^\dag$ & 28.84$^\dag$ \\
&  (i) init + FR(s=3/3) + IT & 1 & 22.9$\times$ & 26.22 & 29.68 & 30.38 & 32.67 \\
&  (j) PLM + FR(s=3/3) + IT & 1& 22.9$\times$& 29.89 & 34.89 & 36.33 & 37.36 \\
&  (k) PLM + DR(s=4/3) + IT & 1 & 22.7$\times$& 29.67 & 34.70 & 36.70 & 37.34 \\
&  (l) PLM + DR(s=4/3) + IM & 1 & 22.7$\times$& 29.86 & 34.88 & 37.66 & 37.91 \\
&  (m) PLM + DR(s=4/3) + IM + ED  & 1 & 22.7$\times$& \bf{29.96} & \bf{34.93} & \bf{37.85} & \bf{38.24} \\
&  (n) PLM + DR(s=4/3) + IM + ED + beam search & 1 & 9.9$\times$& 30.71 & 35.24 & 39.03 & 38.72 \\

\bottomrule
\end{tabular}
\label{tab:main-wmt}
\end{table*}

\paragraph{WMT'14 En$\leftrightarrow$De}
The experimental setup for the WMT'14 En$\leftrightarrow$De dataset is the same as that used for the IWSLT'14 De$\leftrightarrow$En dataset. The results of the experiments are presented in TABLE~\ref{tab:main-wmt}.
\begin{itemize}
    \item For both translation directions: The results of the different settings on the WMT'14 De$\rightarrow$En dataset show a similarity to the results observed on the IWSLT'14 De$\rightarrow$En dataset. 
    Our model surpasses existing NAT models, regardless of whether a pre-trained model is used, in both one-pass and iterative modes. Furthermore, we also outperform Transformer-\textit{base} and Transformer-\textit{big}. Our NAT model sets the state-of-the-art on the WMT14 De$\leftrightarrow$En dataset among NAT models.
    \item In terms of speed, our model achieves a remarkable speedup of 22.7$\times$ compared to the baseline autoregressive model. Even when using beam search, our model still demonstrates a significant speed improvement of 9.9$\times$.    
\end{itemize}
\paragraph{WMT'16 En$\leftrightarrow$Ro}
As noted during the parameter search, the upsampling rate for models initialized with mBERT is found to be s=3.0.
\begin{itemize}
    \item Rows (i), (j), and (k) demonstrate similar trends observed in our previous experiments on the IWSLT'14 De$\rightarrow$En and WMT14 tasks, where initialization with a PMLM and a strong teacher model for distillation contribute to a significant boost in performance.
    \item Moving on to rows (k) and (l), we observe the same effect we have previously seen on IWSLT'14 where IM in place of IT through upsampling improves model performance.
    \item As with WMT'14 En$\leftrightarrow$De, our model surpasses existing NAT models, regardless of whether a pre-trained model is used, in both one-pass and iterative modes. Furthermore, we also outperform Transformer-\textit{base}. Our NAT model sets the state-of-the-art on the WMT'16 En$\leftrightarrow$Ro dataset among NAT models.
\end{itemize}

\paragraph{Various AT Teachers}
Next, let us examine the impact of different KD teachers on CTCPMLM. We use three different types of data: undistilled original data, distillation data from base transformer models, and distillation data from strong teachers. From TABLE~\ref{tab:distill-teacher}, we can observe several phenomena:
\begin{itemize}
    \item Using distillation data from \textit{base} or strong KD teachers results in better performance compared to AT Transformer \textit{base}. Furthermore, even using strong KD teachers outperforms AT Strong Teacher on the IWSLT'14 dataset.
    \item Without utilizing KD, our model outperforms the AT Transformer \textit{base} on both the WMT'16 and IWSLT'14 datasets. Additionally, our model without KD surpasses the current NAT model on the IWSLT'14 dataset, which is a remarkable achievement.
    \item However, CTC beam search does not comprehensively help improve the model's performance. This may be due to the fact that the hyperparameters ($\alpha$, $\beta$) = (0.3, 0.9) are not optimal for this particular setting.
    \item While CTC beam search offers a potential avenue for improvement, our experiments suggest that it may not be universally beneficial. Tuning hyperparameters like the ($\alpha$) and ($\beta$) for specific tasks is crucial for optimal performance. In our case, the current configuration (e.g., ($\alpha$, $\beta$) = (0.3, 0.9) might not be ideal for this particular setting. Our investigation into teacher models for beam search training suggests an intriguing limitation. It appears that when the student model (beam search LM) outperforms the teacher model in terms of performance, as observed in the IWSLT'14 En$\rightarrow$De dataset, the teacher's guidance becomes less effective. This observation raises the possibility that the teacher model's limitations may hinder the student model's further progress. While this finding warrants further investigation, it highlights the need for exploring alternative training strategies or teacher selection methods for beam search LMs, particularly when dealing with student models that exhibit exceptional capabilities.
    
\end{itemize}
\begin{table*}[h]
\centering
\small
\caption{Performance comparison in various AT teachers on the test sets of three corpora. The BLEU score is shown in the table, where on the left side of the "$/$" symbol is the output of CTCPMLM, and on the right side is one with the CTC beam search.}
\begin{tabular}{lcccccc}
\toprule
 \multirow{2}{*}{\textbf{Teacher Models}} & 
 \multicolumn{2}{c}{\textbf{WMT'14}}&
 \multicolumn{2}{c}{\textbf{WMT'16}}&
 \multicolumn{2}{c}{\textbf{IWSLT'14}}  \\
 \ & \textbf{EN-DE} & \textbf{DE-EN} & \textbf{EN-RO} & \textbf{RO-EN} & \textbf{EN-DE} & \textbf{DE-EN} \\
\midrule
\textbf{AT} & & & & & & \\
\midrule
Transformer \textit{base} & 27.48 & 31.39 &  34.26 & 33.87 &  28.47 & 34.85 \\
Strong & 30.98 & 36.30 &  41.61 & 40.76 &  30.36  & 38.22  \\
\midrule
\textbf{NAT-Ours} & & & & & & \\
\midrule

\textit{Base} & 28.81/30.01 & 34.83/35.12 &  35.27/36.31 & 35.75/36.24 &  30.62/30.69 & 38.84/38.52 \\
Strong & 29.96/\bf{30.71} & 34.93/\bf{35.24} &  37.85/\bf{39.03} & 38.24/\bf{38.72} &  31.92/\bf{31.94} & \bf{39.93}/36.57 \\
no-KD & 25.99/20.97 & 27.53/26.20 & 33.45/35.55 & 33.99/34.83 & 29.62/30.55 & 37.85/38.57 \\

\bottomrule
\end{tabular}
\label{tab:distill-teacher}
\end{table*}

\subsection{Ablation Study}

In our ablation study, we will first introduce various technological variations or improvements and apply them individually to the model. By comparing them with the baseline model, we can evaluate the impact of each technique on the model's performance. Next, we will explore the differences in these techniques among different models and compare their performance on those models. Finally, we will discuss the effect of different layers in the ED on the model's performance to determine which layer has the greatest impact on the model.
To ensure more efficiency in our experiments, we conducted the following experiments on the IWSLT'14 De$\rightarrow$En dataset.

\paragraph{Impact of variant techniques} The model incorporates four different techniques (distillation data, IT/IM, PMLM initialization, ED) to enhance its performance. The combined effects of these techniques are compared by examining the BLEU scores in TABLE~\ref{tab:combine_tech}. Based on the experimental results, we obtained the following observations:
\begin{itemize}
\item Importance of KD: In each block, regardless of the employed method, KD consistently provides performance gains.
\item PMLM initialization: Similar to KD, initializing with PMLM yields cumulative improvements in model performance, surpassing even the gains obtained from KD. Furthermore, it is noteworthy that using PMLM alone for initialization outperforms the baseline AT teacher.
\item Effect of IM: IM significantly enhances the performance of CTCPMLM, even without PMLM initialization. This observation presents an intriguing open problem and potential area for future work.
\item Effect of ED: When the model is initialized without PMLM, the mismatch between the model and PMLM's representations causes ED to fail in learning from the PMLM teacher, resulting in worse performance. However, when KD and PMLM initialization are used, incorporating ED improves the model's performance. This consistency is observed across the three datasets and six directions as shown TABLE~\ref{tab:main-iwslt} and TABLE\ref{tab:main-wmt}.
\end{itemize}
Overall, the model with KD, IM, ED, and PMLM initialization achieves the best performance.

\begin{table}[htbp]
    \centering
    \small
    \caption{Ablation on IWSLT14 DE$\rightarrow$EN test set with different combinations of techniques. The default setup directly using BiBERT PMLM to initialize and finetune with CTC loss. ($s=4$ and FR)}
    \begin{tabular}{ccccc}
        \toprule
        KD & Mask & PMLM & ED & BLEU\\
        \midrule
        & & & & 26.46 \\
        & \checkmark & & & 27.50\\
        \checkmark & & & & 32.80 \\
        \checkmark& \checkmark  & & & 33.68\\
        \midrule
        & &\checkmark & &   37.89 \\
        & \checkmark&\checkmark & &   37.85\\
        \checkmark& &\checkmark & &   39.45\\
        \checkmark&\checkmark &\checkmark & &   \textbf{39.68}\\
        
        \midrule
         & & & \checkmark &                   22.70\\        
         &\checkmark & &\checkmark &          21.14 \\         
        \checkmark& & &\checkmark & 32.13\\
        \checkmark&\checkmark& &\checkmark &  \textbf{33.91}\\

        \midrule
        &     &\checkmark &\checkmark & 36.47\\
        & \checkmark &\checkmark & \checkmark& 37.85\\
        \checkmark&&\checkmark &\checkmark & 39.74\\
        \checkmark&\checkmark&\checkmark & \checkmark& \textbf{39.93}\\

    \bottomrule
    \end{tabular}
    \label{tab:combine_tech}
\end{table}

\paragraph{Different Backbone Models} In addition to BiBERT and mBERT PMLM models discussed earlier, we also compared other models such as DistilBERT (\verb|distilbert-base-multilingual-cased|\footnote{\url{https://huggingface.co/distilbert-base-multilingual-cased}}) and XLMR (\verb|xlm-roberta-base|\footnote{\url{https://huggingface.co/xlm-roberta-base}}). While we determined the optimal upsampling rates of 3 for pretrained mBERT and 4 for pretrained BiBERT through preliminary experiments, we did not have the opportunity to conduct similar exploratory experiments for the remaining models. Consequently, we decided to use a common upsampling rate of 3 for all models based on two considerations. 
Firstly, literature on non-autoregressive machine translation commonly employs an upsampling rate of 3\cite{gu-kong-2021-fully,shao-etal-2022-one,shao2022nmla}, indicating its effectiveness as a widely accepted choice. Secondly, by using a consistent upsampling rate, we reduced the number of variables in our experiment, facilitating a more straightforward comparison of model performance. Based on the experimental results presented in TABLE~\ref{tab:diff_pmlm}, the following observations were obtained:
\begin{itemize}
    \item The BiBERT PMLM consistently achieved the highest BLEU score, regardless of whether base or strong teacher distillation was used. This observation can be attributed to the fact that BiBERT was specifically pre-trained on German and English, resulting in a more consistent alignment between tokens in both languages. 
    \item Furthermore, DistilBERT exhibits a speed advantage primarily due to its fewer hidden layers, with only 6 layers. On the other hand, XLMR has the slowest speed. Despite having the same number of layers (12) as mBERT and BiBERT, XLMR has a vocabulary size that is 2.4 times larger than mBERT. As a result, it requires significant computational resources, leading to noticeable decreases in speed.
\end{itemize}  
\begin{table}[h]
\centering
\scriptsize
\caption{Performance comparison between various models on the test set of IWSLT'14 DE$\rightarrow$EN. The latency is measured one sentence per batch and compared with mBERT. The upsampling strategy is $s=3.0$ with DR. 'Parms' means the parameters of the model. 'Num.' indicates the number of the hidden layer. 'Voc.' means the size of the vocabulary.
}
\begin{tabular}{ccccccc}
\toprule

\multirow{2}{*}{Models} & \multicolumn{2}{c}{Distillatioin} & \multirow{2}{*}{Speedup} & \multirow{2}{*}{Parms.} & \multirow{2}{*}{Num.} & \multirow{2}{*}{Voc.}\\
& \textit{base} & \textit{strong}\\
\midrule
init(mBERT) & 31.35 & 33.07 & 1.00$\times$ & 167M & 12 & 105,879\\
mBERT & 36.87 & 38.56    & 1.00$\times$ & 167M & 12 & 105,879 \\
BiBERT & \textbf{38.74} & \textbf{39.99} & 1.04$\times$ & 126M & 12 & 52,000 \\
DistilBERT & 34.38 & 35.89 & \textbf{1.48}$\times$ & 135M & 6 & 119,547 \\
XLMR & 36.78 & 38.20 & 0.85$\times$ & 278M & 12 & 250,002 \\
\bottomrule
\end{tabular}

\label{tab:diff_pmlm}
\end{table}

\begin{figure}[h]
    \centering
    \includegraphics[width=0.45\textwidth]{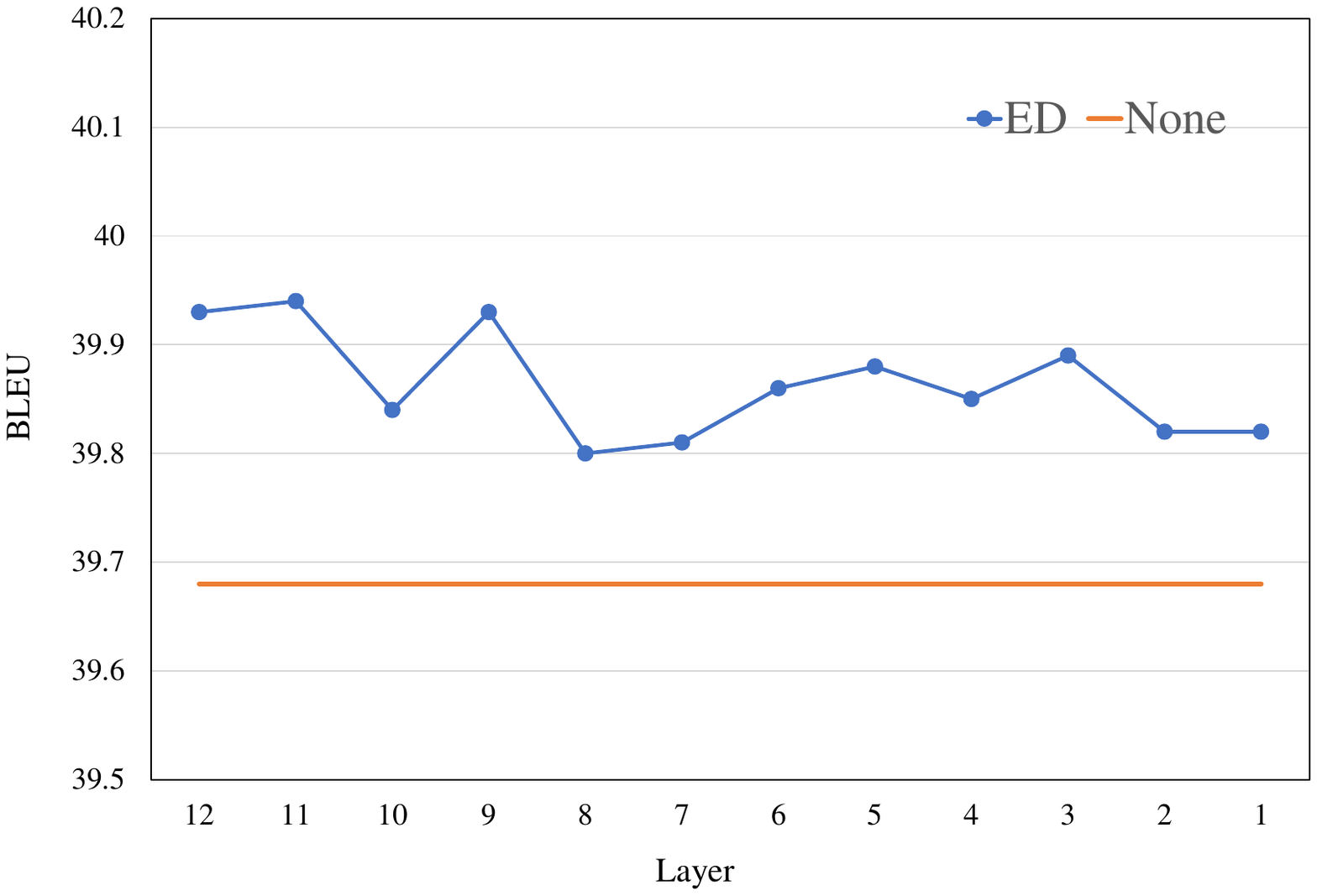}
    \caption{BLEU v.s. Distillation Layer on the test set of IWSLT'14 DE$\rightarrow$EN. We compare different distillation layer on BiBERT. Here "None" means not using ED method.}
    \label{fig:Pretrained Model and KD-M Layer}
\end{figure}

\paragraph{ED with Different Distillation Layers}
In our ED technique, we primarily employ distillation using the last layer of the teacher PMLM. Nevertheless, we conducted additional investigations to explore the impact of distilling from other layers, and we present the corresponding performance on the test set of IWSLT'14 DE$\rightarrow$EN, as shown in Figure \ref{fig:Pretrained Model and KD-M Layer}. From the experimental results, the following observations were made:
\begin{itemize}
  \item The utilization of the ED technique consistently enhances the BLEU score, irrespective of the layer being employed.
  \item The model implementing the ED technique displays a consistent range of BLEU scores across various hidden layers, spanning from 39.8 to 39.94.
  \item Among these findings, the model featuring the last 2 layers attains the highest BLEU score, with a diminishing trend observed as we move toward the preceding layers.
\end{itemize}

Based on the data, it can be concluded that the utilization of ED techniques positively influences model performance, particularly in the case of the last few hidden layers, resulting in the highest BLEU scores.

\section{Related work utilizing PLM initialization In NAT} \label{plm_relate_dwork}
NAT approaches typically involve additional mechanisms alongside PLMs. AB-Net~\cite{10.5555/3495724.3496634} employs adapter layers within each BERT layer and incorporates a special [LENGTH] token in the encoder to predict sequence length. Bert+CRF-NAT~\cite{su-etal-2021-non} leverages BERT with a Conditional Random Field (CRF) to improve the modeling of output-side dependencies. It employs padding tokens to reach a predefined maximum sequence length and utilizes [eos] tokens for end-of-sentence prediction. DEER~\cite{liang-etal-2023-dynamic} initializes with XLM-R and introduces the Levenshtein Transformer~\cite{gu2019levenshteintransformer} for handling insertion, deletion, and placeholder operations. XLM-D~\cite{wang-etal-2022-xlm}, initialized with XLM-R, incorporates a distance-based latent transformation module and a position-wise add and scale module to ensure consistency within the model's intermediate representation.
Our approach deviates from these methods by directly utilizing PLMs without integrating adapters into any layer. This approach preserves the inherent capabilities of the PLM across all layers and reduces the computational overhead associated with inference and storage. Additionally, we leverage CTC, which eliminates the need for explicit sequence length prediction, mitigating the influence of length prediction accuracy on overall performance~\cite{Gu0XLS18}. Furthermore, to further preserve the rich knowledge embedded within the PLM, we adopt the ED technique to maintain knowledge from the original PLM.

\section{Discussion} \label{discussion}
\subsection{Does the dependency on KD become looser as the model becomes stronger ?}
In TABLE~\ref{tab:combine_tech}, our model achieves 37.85 without KD using only PMLM initialization. Surprisingly, this outperforms NAT models with KD listed in TABLE~\ref{tab:main-iwslt}. This result shows that our model can be very effective even without KD and can outperform some NAT models that use KD. Further, when we use PMLM initialization with stronger KD, the BLEU score increases from 37.85 to 39.93. Therefore, based on our findings, we suggest that stronger knowledge distillation (KD) can indeed lead to improved performance for a stronger model. However, this does not imply a complete disregard for the role of KD in the training process. Knowledge distillation still remains an important component in enhancing the performance of the model.
\subsection{How about the efficiency when utilizing ED ?}
\paragraph{Complexity}
In CTC Loss, the training cost comes from source length $\vert x \vert$, hidden dimension $d$ and upsampling rate $s$, and the complexity per layer is $O(\vert x \vert^2s^2d)$\cite{keles2022computational}. 
In the ED Loss, after calculating the ED loss matrix denoted as $Q_{ij}=\log p(y_i\vert \hat{h}_{nat_j})$, we use the Hungarian algorithm\cite{https://doi.org/10.1002/nav.3800020109,pmlr-v139-du21c} to find the optimal alignment based on the probability matrix $Q$. The complexity of the Hungarian algorithm is $O(\vert x \vert^2s^2|y|)$\cite{Jonker1987ASA}. The total complexity can be approximated as $O(\vert x \vert^2s^2d\times l) + O(\vert x \vert^2s^2|y|)$, where $l$ is the number of layers, and $|y|\ll d\times l$. Therefore, the total complexity is primarily determined by computing the CTC loss, making the computation of the ED loss negligible or, at least, less impactful.

\paragraph{Actual Training Time}
TABLE~\ref{tab:ed-training} presents a comparison of training time consumption between using ED and not using ED, indicating that utilizing ED results in a $1.72 \text{--} 4.69\%$ increase. \\
Finally, considering the complexity and actual training time, we can know that using the ED method will not bring a significant burden to the training process. Furthermore, during inference, the model does not utilize the ED process, resulting in no additional time cost.
\begin{table}[htbp]
\centering
\scriptsize
\caption{A comparison of training time consumption between using ED and not using ED. The upsampling rate of $s=4$. ’Step’ means the number of training steps. ’Batch’ means number of tokens in a batch. ’Time’ means the training time measured on 4 GeForce RTX 3090 GPUs. The '\%' symbol represents the percentage increase resulting from utilizing ED
}
\begin{tabular}{lccccc}
\toprule
Datasets  & Step & Batch & w/ED & w/o ED & \% \\
\midrule
IWSLT'14 De$\rightarrow$En  & 50K & 12288 & 5.9hrs &  5.8hrs & 1.72\% \\
WMT'14 De$\rightarrow$En  & 100K & 66536 & 43.5hrs &  41.60hrs & 4.56\% \\
WMT'16 Ro$\rightarrow$En  & 30K & 65536 & 15.4hrs &  14.71hrs & 4.69\% \\

\bottomrule
\end{tabular}
\label{tab:ed-training}
\end{table}

\subsection{Can Vocabulary Pruning be utilized to further enhance speed while maintaining performance?}
The previous studies by \cite{DBLP:journals/corr/abs-1910-10683, martins2020sparse} delve into vocabulary selection for transformer-based models and showcase that substantial reductions in vocabulary size can be accomplished without compromising model performance. This reduction in vocabulary size not only enhances memory efficiency but also accelerates the model's inference time. In our research, we employ TextPruner \cite{yang-etal-2022-textpruner} to prune the vocabulary. The result is in TABLE~\ref{tab:voc_pruning}. From the data, it can be observed that using pruned-mBERT significantly reduces the vocabulary size. This reduction is because mBERT is pretrained on 102 languages. In contrast, BiBERT does not exhibit this phenomenon. Furthermore, using pruned-mBERT improves performance and speeds up inference.

\begin{table}[h]
\centering
\small
\caption{Performance comparison of vocabulary pruning models on the test set of IWSLT'14 DE$\rightarrow$EN. Latency is measured with one sentence per batch and compared to using mBERT and BiBERT initialization. "Voc. Size" refers to the vocabulary size.}
\begin{tabular}{ccccc}
\toprule
Model & Voc. Size & Parms. & Speedup & BLEU\\
\midrule
mBERT & 105,879 & 167M & 1.00$\times$ & 38.15 \\
pruned-mBERT & 26,458 & 106M & \bf{1.11}$\times$ & \bf{38.54} \\
\midrule
BiBERT & 52,000 & 126M & 1.00$\times$ & 39.93 \\
pruned-BiBERT & 43,093 & 119M & 1.01$\times$ & 39.51 \\
\bottomrule
\end{tabular}
\label{tab:voc_pruning}
\end{table}

\subsection{Results on Low-Resource Language Pairs}
In addition, we evaluate the efficacy of our framework on three low-resource language pairs from the IWSLT'14 dataset (English$\leftrightarrow$ Italian/Spanish/Dutch (IWSLT14 En$\leftrightarrow$ It/Es/Nl). As shown in TABLE~\ref{tab:low_resource}, the results for both translation directions, measured using BLEU score, indicate that the proposed CTCPMLM consistently surpasses the performance of the baseline models, thereby demonstrating the robustness and generality of our approach.

\begin{table}[H]
\centering
\color{black}
\scriptsize
\caption{
 The performance on IWSLT'14 low-resource language pairs.$^\Diamond$ denotes results from our run
}
\begin{tabular}{lcccccc}
\toprule
 Model & \textbf{EN-IT} & \textbf{IT-EN} & \textbf{EN-ES} & \textbf{ES-EN} & \textbf{EN-NL} & \textbf{NL-EN}\\
\midrule
Transformer \textit{base} & 29.50 & 31.91 & 36.88 & 40.47 & 31.96 & 37.00 \\
AB-Net                   & \textbf{31.81} &   34.20    & 37.45 &   42.66   &  32.52     & 38.94 \\
CTCPMLM                   & 31.48 &  \textbf{34.91}  & \textbf{39.02} &   \textbf{43.44}    & \textbf{33.74} & \textbf{39.86} \\

\bottomrule
\end{tabular}
\label{tab:low_resource}
\end{table}

\subsection{Is the performance of the model due to the increase in the number of parameters?}
In this section, we investigate whether the improved performance is solely due to the increased number of model parameters. An analysis of their performance on the WMT'14 En$\rightarrow$ De as detailed in Table~\ref{tab:para_vs_bleu}, revealed that CTCPMLM consistently achieved the highest BLEU score across all parameter sizes (44M, 85M, and 176M) in both NAT and AT settings. This finding underscores that CTCPMLM success is not solely attributable to parameter size.



\begin{table}[h]
\centering
\caption{Comparing Model Parameter Size and Performance on WMT'14 En$\rightarrow$De Test Set.The BLEU score is shown in the table.$^*$ denotes models trained with distillation from a \textit{transformer big},emb. denotes embedding, wt. denotes weighting}
\label{tab:para_vs_bleu}
\color{black}
\scriptsize
\begin{tabular}{lcccccc}
\toprule
\multirow{2}{*}{\textbf{Models}} &\multirow{2}{*}{\textbf{Iter.}} & \multirow{2}{*}{\textbf{Speed}}  & \multicolumn{2}{c}{\textbf{Parameters}} &  & \multirow{2}{*}{\textbf{BLEU}} \\
\cline{4-5} &  &  & \textbf{emb.+wt.} & \textbf{wt.}  &  & \\
\midrule
Transformer-\textit{base}         & N & 1$\times$   & 65M  & 44M  &  & 27.48\\
Transformer-\textit{big}          & N & -           & 218M & 176M &  & 28.40\\
XLM-D~\cite{wang-etal-2022-xlm}   & 1 & 19.5$\times$ & -    & 85M & & 26.91\\
DEER~\cite{liang-etal-2023-dynamic} & 1 & 12.0$\times$ & -    & 85M & & 26.19 \\
Fully-NAT(\textit{base})~\cite{gu-kong-2021-fully} & 1 & 16.5$\times$ & -    & 44M & & 27.49 \\
Fully-NAT(\textit{big})~\cite{gu-kong-2021-fully} & 1 & 15.8$\times$ & -    & 176M & & 27.89$^*$ \\
CTCPMLM                           & 1 & 22.7$\times$& 126M & 85M & & \textbf{28.81}  \\
\bottomrule
\end{tabular}
\end{table}


\begin{table}[h]
\centering
\caption{Pruning the fine-tuned CTCPMLM on the IWSLT'14 En$\rightarrow$De Test Set. The BLEU score is shown in the table.}
\label{tab:iwslt14_pruning}
\begin{tabular}{cc}
\toprule
\textbf{Percentage} & \textbf{BLEU} \\
\midrule
100\% & 39.93 \\
90\% & 39.91 \\
80\% & 39.82 \\
70\% & 39.27 \\
60\% & 37.82 \\
\bottomrule
\end{tabular}
\end{table}

\section{Conclusion} \label{conclusion}

This paper explores the potential of using PMLM with CTC loss for NAT models on three translation datasets. To overcome the position information loss in CTC upsampling, we propose a dynamic CTC upsampling strategy to determine the upsampling ratio. Additionally, we find that the "Inserting Masks" (IM) strategy works better for CTCPMLM than the "Inserting Tokens" (IT) strategy. We introduce the ED method, which allows the NAT model to distill contextualized information from the pre-trained PMLM in the target space.
In the experiments, we demonstrate that our methods achieve state-of-the-art results on the IWSLT'14 De$\leftrightarrow$En translation task. Furthermore, our approach outperforms all existing NAT models on the WMT'14 En$\leftrightarrow$De and WMT'16 En$\leftrightarrow$Ro datasets, and even shows competitive performance with many strong auto-regressive models.

\section{Acknowledgements} \label{ack}
This work was supported by the National Science and Technology Council,
R.O.C. Taiwan under Grants 112-2628-E-002-026 and 110-2223-E-002-007-MY3.

\bibliographystyle{IEEEtran}
\bibliography{IEEEtran, custom, anthology}

\begin{thebibliography}{10}
\providecommand{\url}[1]{#1}
\csname url@samestyle\endcsname
\providecommand{\newblock}{\relax}
\providecommand{\bibinfo}[2]{#2}
\providecommand{\BIBentrySTDinterwordspacing}{\spaceskip=0pt\relax}
\providecommand{\BIBentryALTinterwordstretchfactor}{4}
\providecommand{\BIBentryALTinterwordspacing}{\spaceskip=\fontdimen2\font plus
\BIBentryALTinterwordstretchfactor\fontdimen3\font minus \fontdimen4\font\relax}
\providecommand{\BIBforeignlanguage}[2]{{%
\expandafter\ifx\csname l@#1\endcsname\relax
\typeout{** WARNING: IEEEtran.bst: No hyphenation pattern has been}%
\typeout{** loaded for the language `#1'. Using the pattern for}%
\typeout{** the default language instead.}%
\else
\language=\csname l@#1\endcsname
\fi
#2}}
\providecommand{\BIBdecl}{\relax}
\BIBdecl

\bibitem{mikolov2010recurrent}
T.~Mikolov, M.~Karafi{\'a}t, L.~Burget, J.~Cernock{\`y}, and S.~Khudanpur, ``Recurrent neural network based language model.'' in \emph{Interspeech}, vol.~2, no.~3.\hskip 1em plus 0.5em minus 0.4em\relax Makuhari, 2010, pp. 1045--1048.

\bibitem{bahdanau2014neural}
D.~Bahdanau, K.~Cho, and Y.~Bengio, ``Neural machine translation by jointly learning to align and translate,'' \emph{arXiv preprint arXiv:1409.0473}, 2014.

\bibitem{vaswani2017attention}
A.~Vaswani, N.~Shazeer, N.~Parmar, J.~Uszkoreit, L.~Jones, A.~N. Gomez, {\L}.~Kaiser, and I.~Polosukhin, ``Attention is all you need,'' in \emph{NIPS}, 2017, pp. 5998--6008.

\bibitem{Gu0XLS18}
\BIBentryALTinterwordspacing
J.~Gu, J.~Bradbury, C.~Xiong, V.~O.~K. Li, and R.~Socher, ``Non-autoregressive neural machine translation,'' in \emph{Proceedings of the 6th International Conference on Learning Representations ({ICLR})}.\hskip 1em plus 0.5em minus 0.4em\relax Vancouver, BC, Canada: OpenReview.net, May 2018, pp. 1--13. [Online]. Available: \url{https://openreview.net/forum?id=B1l8BtlCb}
\BIBentrySTDinterwordspacing

\bibitem{stern19a}
\BIBentryALTinterwordspacing
M.~Stern, W.~Chan, J.~Kiros, and J.~Uszkoreit, ``Insertion transformer: Flexible sequence generation via insertion operations,'' in \emph{Proceedings of the 36th International Conference on Machine Learning}, ser. Proceedings of Machine Learning Research, K.~Chaudhuri and R.~Salakhutdinov, Eds., vol.~97.\hskip 1em plus 0.5em minus 0.4em\relax Long Beach, California, USA: PMLR, 09--15 Jun 2019, pp. 5976--5985. [Online]. Available: \url{http://proceedings.mlr.press/v97/stern19a.html}
\BIBentrySTDinterwordspacing

\bibitem{GuWZ19}
\BIBentryALTinterwordspacing
J.~Gu, C.~Wang, and J.~Zhao, ``Levenshtein transformer,'' in \emph{Advances in Neural Information Processing Systems 32: Annual Conference on Neural Information Processing Systems (NeurIPS)}, H.~M. Wallach, H.~Larochelle, A.~Beygelzimer, F.~d'Alch{\'{e}}{-}Buc, E.~B. Fox, and R.~Garnett, Eds., Vancouver, BC, Canada, 2019, pp. 11\,179--11\,189. [Online]. Available: \url{https://proceedings.neurips.cc/paper/2019/hash/675f9820626f5bc0afb47b57890b466e-Abstract.html}
\BIBentrySTDinterwordspacing

\bibitem{shu2020latent}
R.~Shu, J.~Lee, H.~Nakayama, and K.~Cho, ``Latent-variable non-autoregressive neural machine translation with deterministic inference using a delta posterior,'' in \emph{AAAI}, vol.~34, no.~05, 2020, pp. 8846--8853.

\bibitem{libovicky-helcl-2018-end}
\BIBentryALTinterwordspacing
J.~Libovick{\'y} and J.~Helcl, ``End-to-end non-autoregressive neural machine translation with connectionist temporal classification,'' in \emph{Proceedings of the 2018 Conference on Empirical Methods in Natural Language Processing}.\hskip 1em plus 0.5em minus 0.4em\relax Brussels, Belgium: Association for Computational Linguistics, Oct.-Nov. 2018, pp. 3016--3021. [Online]. Available: \url{https://aclanthology.org/D18-1336}
\BIBentrySTDinterwordspacing

\bibitem{saharia-etal-2020-non}
\BIBentryALTinterwordspacing
C.~Saharia, W.~Chan, S.~Saxena, and M.~Norouzi, ``Non-autoregressive machine translation with latent alignments,'' in \emph{Proceedings of the 2020 Conference on Empirical Methods in Natural Language Processing (EMNLP)}.\hskip 1em plus 0.5em minus 0.4em\relax Online: Association for Computational Linguistics, Nov. 2020, pp. 1098--1108. [Online]. Available: \url{https://aclanthology.org/2020.emnlp-main.83}
\BIBentrySTDinterwordspacing

\bibitem{gu-kong-2021-fully}
\BIBentryALTinterwordspacing
J.~Gu and X.~Kong, ``Fully non-autoregressive neural machine translation: Tricks of the trade,'' in \emph{Findings of the Association for Computational Linguistics: ACL-IJCNLP 2021}.\hskip 1em plus 0.5em minus 0.4em\relax Online: Association for Computational Linguistics, Aug. 2021, pp. 120--133. [Online]. Available: \url{https://aclanthology.org/2021.findings-acl.11}
\BIBentrySTDinterwordspacing

\bibitem{shao-etal-2022-one}
\BIBentryALTinterwordspacing
C.~Shao, X.~Wu, and Y.~Feng, ``One reference is not enough: Diverse distillation with reference selection for non-autoregressive translation,'' in \emph{Proceedings of the 2022 Conference of the North American Chapter of the Association for Computational Linguistics: Human Language Technologies}.\hskip 1em plus 0.5em minus 0.4em\relax Seattle, United States: Association for Computational Linguistics, Jul. 2022, pp. 3779--3791. [Online]. Available: \url{https://aclanthology.org/2022.naacl-main.277}
\BIBentrySTDinterwordspacing

\bibitem{shao2022nmla}
\BIBentryALTinterwordspacing
C.~Shao and Y.~Feng, ``Non-monotonic latent alignments for ctc-based non-autoregressive machine translation,'' 2022. [Online]. Available: \url{https://arxiv.org/abs/2210.03953}
\BIBentrySTDinterwordspacing

\bibitem{wang-etal-2022-xlm}
\BIBentryALTinterwordspacing
Y.~Wang, S.~He, G.~Chen, Y.~Chen, and D.~Jiang, ``{XLM}-{D}: Decorate cross-lingual pre-training model as non-autoregressive neural machine translation,'' in \emph{Proceedings of the 2022 Conference on Empirical Methods in Natural Language Processing}, Y.~Goldberg, Z.~Kozareva, and Y.~Zhang, Eds.\hskip 1em plus 0.5em minus 0.4em\relax Abu Dhabi, United Arab Emirates: Association for Computational Linguistics, Dec. 2022, pp. 6934--6946. [Online]. Available: \url{https://aclanthology.org/2022.emnlp-main.466}
\BIBentrySTDinterwordspacing

\bibitem{liang-etal-2023-dynamic}
\BIBentryALTinterwordspacing
X.~Liang, J.~Li, L.~Wu, Z.~Cao, and M.~Zhang, ``Dynamic and efficient inference for text generation via {BERT} family,'' in \emph{Proceedings of the 61st Annual Meeting of the Association for Computational Linguistics (Volume 1: Long Papers)}, A.~Rogers, J.~Boyd-Graber, and N.~Okazaki, Eds.\hskip 1em plus 0.5em minus 0.4em\relax Toronto, Canada: Association for Computational Linguistics, Jul. 2023, pp. 2883--2897. [Online]. Available: \url{https://aclanthology.org/2023.acl-long.162}
\BIBentrySTDinterwordspacing

\bibitem{GravesFGS06}
\BIBentryALTinterwordspacing
A.~Graves, S.~Fern{\'{a}}ndez, F.~J. Gomez, and J.~Schmidhuber, ``Connectionist temporal classification: labelling unsegmented sequence data with recurrent neural networks,'' in \emph{Proceedings of the 23rd International Conference on Machine Learning ({ICML})}, W.~W. Cohen and A.~W. Moore, Eds.\hskip 1em plus 0.5em minus 0.4em\relax Pittsburgh, PA, USA: {ACM}, 2006, pp. 369--376. [Online]. Available: \url{https://doi.org/10.1145/1143844.1143891}
\BIBentrySTDinterwordspacing

\bibitem{zhou2021understanding}
C.~Zhou, G.~Neubig, and J.~Gu, ``Understanding knowledge distillation in non-autoregressive machine translation,'' 2021.

\bibitem{kim-rush-2016-sequence}
\BIBentryALTinterwordspacing
Y.~Kim and A.~M. Rush, ``Sequence-level knowledge distillation,'' in \emph{Proceedings of the 2016 Conference on Empirical Methods in Natural Language Processing}.\hskip 1em plus 0.5em minus 0.4em\relax Austin, Texas: Association for Computational Linguistics, Nov. 2016, pp. 1317--1327. [Online]. Available: \url{https://aclanthology.org/D16-1139}
\BIBentrySTDinterwordspacing

\bibitem{lee-etal-2018-deterministic}
\BIBentryALTinterwordspacing
J.~Lee, E.~Mansimov, and K.~Cho, ``Deterministic non-autoregressive neural sequence modeling by iterative refinement,'' in \emph{Proceedings of the 2018 Conference on Empirical Methods in Natural Language Processing}.\hskip 1em plus 0.5em minus 0.4em\relax Brussels, Belgium: Association for Computational Linguistics, Oct.-Nov. 2018, pp. 1173--1182. [Online]. Available: \url{https://aclanthology.org/D18-1149}
\BIBentrySTDinterwordspacing

\bibitem{ghazvininejad-etal-2019-mask}
\BIBentryALTinterwordspacing
M.~Ghazvininejad, O.~Levy, Y.~Liu, and L.~Zettlemoyer, ``Mask-predict: Parallel decoding of conditional masked language models,'' in \emph{Proceedings of the 2019 Conference on Empirical Methods in Natural Language Processing and the 9th International Joint Conference on Natural Language Processing (EMNLP-IJCNLP)}.\hskip 1em plus 0.5em minus 0.4em\relax Hong Kong, China: Association for Computational Linguistics, Nov. 2019, pp. 6112--6121. [Online]. Available: \url{https://aclanthology.org/D19-1633}
\BIBentrySTDinterwordspacing

\bibitem{xiao2023amomadaptivemaskingmasking}
\BIBentryALTinterwordspacing
Y.~Xiao, R.~Xu, L.~Wu, J.~Li, T.~Qin, Y.-T. Liu, and M.~Zhang, ``Amom: Adaptive masking over masking for conditional masked language model,'' 2023. [Online]. Available: \url{https://arxiv.org/abs/2303.07457}
\BIBentrySTDinterwordspacing

\bibitem{lample2019cross}
G.~Lample and A.~Conneau, ``Cross-lingual language model pretraining,'' \emph{arXiv preprint arXiv:1901.07291}, 2019.

\bibitem{zanon-boito-etal-2020-mass}
\BIBentryALTinterwordspacing
M.~Zanon~Boito, W.~Havard, M.~Garnerin, {\'E}.~Le~Ferrand, and L.~Besacier, ``\BIBforeignlanguage{English}{{M}a{SS}: A large and clean multilingual corpus of sentence-aligned spoken utterances extracted from the {B}ible},'' in \emph{\BIBforeignlanguage{English}{Proceedings of the Twelfth Language Resources and Evaluation Conference}}.\hskip 1em plus 0.5em minus 0.4em\relax Marseille, France: European Language Resources Association, May 2020, pp. 6486--6493. [Online]. Available: \url{https://aclanthology.org/2020.lrec-1.799}
\BIBentrySTDinterwordspacing

\bibitem{liu-etal-2020-multilingual-denoising}
\BIBentryALTinterwordspacing
Y.~Liu, J.~Gu, N.~Goyal, X.~Li, S.~Edunov, M.~Ghazvininejad, M.~Lewis, and L.~Zettlemoyer, ``Multilingual denoising pre-training for neural machine translation,'' \emph{Transactions of the Association for Computational Linguistics}, vol.~8, pp. 726--742, 2020. [Online]. Available: \url{https://aclanthology.org/2020.tacl-1.47}
\BIBentrySTDinterwordspacing

\bibitem{xue-etal-2021-mt5}
\BIBentryALTinterwordspacing
L.~Xue, N.~Constant, A.~Roberts, M.~Kale, R.~Al-Rfou, A.~Siddhant, A.~Barua, and C.~Raffel, ``m{T}5: A massively multilingual pre-trained text-to-text transformer,'' in \emph{Proceedings of the 2021 Conference of the North American Chapter of the Association for Computational Linguistics: Human Language Technologies}.\hskip 1em plus 0.5em minus 0.4em\relax Online: Association for Computational Linguistics, Jun. 2021, pp. 483--498. [Online]. Available: \url{https://aclanthology.org/2021.naacl-main.41}
\BIBentrySTDinterwordspacing

\bibitem{li-etal-2022-universal}
\BIBentryALTinterwordspacing
P.~Li, L.~Li, M.~Zhang, M.~Wu, and Q.~Liu, ``Universal conditional masked language pre-training for neural machine translation,'' in \emph{Proceedings of the 60th Annual Meeting of the Association for Computational Linguistics (Volume 1: Long Papers)}.\hskip 1em plus 0.5em minus 0.4em\relax Dublin, Ireland: Association for Computational Linguistics, May 2022, pp. 6379--6391. [Online]. Available: \url{https://aclanthology.org/2022.acl-long.442}
\BIBentrySTDinterwordspacing

\bibitem{rothe-etal-2020-leveraging}
\BIBentryALTinterwordspacing
S.~Rothe, S.~Narayan, and A.~Severyn, ``Leveraging pre-trained checkpoints for sequence generation tasks,'' \emph{Transactions of the Association for Computational Linguistics}, vol.~8, pp. 264--280, 2020. [Online]. Available: \url{https://aclanthology.org/2020.tacl-1.18}
\BIBentrySTDinterwordspacing

\bibitem{Zhu2020Incorporating}
J.~Zhu, Y.~Xia, L.~Wu, D.~He, T.~Qin, W.~Zhou, H.~Li, and T.-Y. Liu, ``Incorporating bert into neural machine translation,'' \emph{arXiv preprint arXiv:2002.06823}, 2020.

\bibitem{10.5555/3495724.3496634}
J.~Guo, Z.~Zhang, L.~Xu, H.-R. Wei, B.~Chen, and E.~Chen, ``Incorporating bert into parallel sequence decoding with adapters,'' in \emph{Proceedings of the 34th International Conference on Neural Information Processing Systems}, ser. NIPS'20.\hskip 1em plus 0.5em minus 0.4em\relax Red Hook, NY, USA: Curran Associates Inc., 2020.

\bibitem{su-etal-2021-non}
\BIBentryALTinterwordspacing
Y.~Su, D.~Cai, Y.~Wang, D.~Vandyke, S.~Baker, P.~Li, and N.~Collier, ``Non-autoregressive text generation with pre-trained language models,'' in \emph{Proceedings of the 16th Conference of the European Chapter of the Association for Computational Linguistics: Main Volume}.\hskip 1em plus 0.5em minus 0.4em\relax Online: Association for Computational Linguistics, Apr. 2021, pp. 234--243. [Online]. Available: \url{https://aclanthology.org/2021.eacl-main.18}
\BIBentrySTDinterwordspacing

\bibitem{VaswaniSPUJGKP17}
\BIBentryALTinterwordspacing
A.~Vaswani, N.~Shazeer, N.~Parmar, J.~Uszkoreit, L.~Jones, A.~N. Gomez, L.~Kaiser, and I.~Polosukhin, ``Attention is all you need,'' in \emph{Advances in Neural Information Processing Systems 30: Annual Conference on Neural Information Processing Systems (NeurIPS)}, I.~Guyon, U.~von Luxburg, S.~Bengio, H.~M. Wallach, R.~Fergus, S.~V.~N. Vishwanathan, and R.~Garnett, Eds., Long Beach, CA, USA, 2017, pp. 5998--6008. [Online]. Available: \url{https://proceedings.neurips.cc/paper/2017/hash/3f5ee243547dee91fbd053c1c4a845aa-Abstract.html}
\BIBentrySTDinterwordspacing

\bibitem{https://doi.org/10.48550/arxiv.2204.09269}
\BIBentryALTinterwordspacing
Y.~Xiao, L.~Wu, J.~Guo, J.~Li, M.~Zhang, T.~Qin, and T.-y. Liu, ``A survey on non-autoregressive generation for neural machine translation and beyond,'' 2022. [Online]. Available: \url{https://arxiv.org/abs/2204.09269}
\BIBentrySTDinterwordspacing

\bibitem{GravesCTC06}
\BIBentryALTinterwordspacing
A.~Graves, S.~Fern\'{a}ndez, F.~Gomez, and J.~Schmidhuber, ``Connectionist temporal classification: Labelling unsegmented sequence data with recurrent neural networks,'' in \emph{Proceedings of the 23rd International Conference on Machine Learning}, ser. ICML '06.\hskip 1em plus 0.5em minus 0.4em\relax New York, NY, USA: Association for Computing Machinery, 2006, p. 369–376. [Online]. Available: \url{https://doi.org/10.1145/1143844.1143891}
\BIBentrySTDinterwordspacing

\bibitem{Graves2013CTC}
\BIBentryALTinterwordspacing
A.~Graves, A.-r. Mohamed, and G.~Hinton, ``Speech recognition with deep recurrent neural networks,'' 2013. [Online]. Available: \url{https://arxiv.org/abs/1303.5778}
\BIBentrySTDinterwordspacing

\bibitem{wei-etal-2019-imitation}
\BIBentryALTinterwordspacing
B.~Wei, M.~Wang, H.~Zhou, J.~Lin, and X.~Sun, ``Imitation learning for non-autoregressive neural machine translation,'' in \emph{Proceedings of the 57th Annual Meeting of the Association for Computational Linguistics}.\hskip 1em plus 0.5em minus 0.4em\relax Florence, Italy: Association for Computational Linguistics, Jul. 2019, pp. 1304--1312. [Online]. Available: \url{https://aclanthology.org/P19-1125}
\BIBentrySTDinterwordspacing

\bibitem{graves2006connectionist}
A.~Graves, S.~Fern{\'a}ndez, F.~Gomez, and J.~Schmidhuber, ``Connectionist temporal classification: labelling unsegmented sequence data with recurrent neural networks,'' in \emph{ICML}, 2006, pp. 369--376.

\bibitem{chan2020imputer}
W.~Chan, C.~Saharia, G.~Hinton, M.~Norouzi, and N.~Jaitly, ``Imputer: Sequence modelling via imputation and dynamic programming,'' in \emph{ICML}.\hskip 1em plus 0.5em minus 0.4em\relax PMLR, 2020, pp. 1403--1413.

\bibitem{https://doi.org/10.1002/nav.3800020109}
\BIBentryALTinterwordspacing
H.~W. Kuhn, ``The hungarian method for the assignment problem,'' \emph{Naval Research Logistics Quarterly}, vol.~2, no. 1-2, pp. 83--97, 1955. [Online]. Available: \url{https://onlinelibrary.wiley.com/doi/abs/10.1002/nav.3800020109}
\BIBentrySTDinterwordspacing

\bibitem{pmlr-v139-du21c}
\BIBentryALTinterwordspacing
C.~Du, Z.~Tu, and J.~Jiang, ``Order-agnostic cross entropy for non-autoregressive machine translation,'' in \emph{Proceedings of the 38th International Conference on Machine Learning}, ser. Proceedings of Machine Learning Research, M.~Meila and T.~Zhang, Eds., vol. 139.\hskip 1em plus 0.5em minus 0.4em\relax PMLR, 18--24 Jul 2021, pp. 2849--2859. [Online]. Available: \url{https://proceedings.mlr.press/v139/du21c.html}
\BIBentrySTDinterwordspacing

\bibitem{xu-etal-2021-bert}
\BIBentryALTinterwordspacing
H.~Xu, B.~Van~Durme, and K.~Murray, ``{BERT}, m{BERT}, or {B}i{BERT}? a study on contextualized embeddings for neural machine translation,'' in \emph{Proceedings of the 2021 Conference on Empirical Methods in Natural Language Processing}.\hskip 1em plus 0.5em minus 0.4em\relax Online and Punta Cana, Dominican Republic: Association for Computational Linguistics, Nov. 2021, pp. 6663--6675. [Online]. Available: \url{https://aclanthology.org/2021.emnlp-main.534}
\BIBentrySTDinterwordspacing

\bibitem{cettolo-etal-2014-report}
\BIBentryALTinterwordspacing
M.~Cettolo, J.~Niehues, S.~St{\"u}ker, L.~Bentivogli, and M.~Federico, ``Report on the 11th {IWSLT} evaluation campaign,'' in \emph{Proceedings of the 11th International Workshop on Spoken Language Translation: Evaluation Campaign}, Lake Tahoe, California, Dec. 4-5 2014, pp. 2--17. [Online]. Available: \url{https://aclanthology.org/2014.iwslt-evaluation.1}
\BIBentrySTDinterwordspacing

\bibitem{bojar-etal-2016-findings}
\BIBentryALTinterwordspacing
O.~Bojar, R.~Chatterjee, C.~Federmann, Y.~Graham, B.~Haddow, M.~Huck, A.~Jimeno~Yepes, P.~Koehn, V.~Logacheva, C.~Monz, M.~Negri, A.~N{\'e}v{\'e}ol, M.~Neves, M.~Popel, M.~Post, R.~Rubino, C.~Scarton, L.~Specia, M.~Turchi, K.~Verspoor, and M.~Zampieri, ``Findings of the 2016 conference on machine translation,'' in \emph{Proceedings of the First Conference on Machine Translation: Volume 2, Shared Task Papers}.\hskip 1em plus 0.5em minus 0.4em\relax Berlin, Germany: Association for Computational Linguistics, Aug. 2016, pp. 131--198. [Online]. Available: \url{https://aclanthology.org/W16-2301}
\BIBentrySTDinterwordspacing

\bibitem{bojar-etal-2014-findings}
\BIBentryALTinterwordspacing
O.~Bojar, C.~Buck, C.~Federmann, B.~Haddow, P.~Koehn, J.~Leveling, C.~Monz, P.~Pecina, M.~Post, H.~Saint-Amand, R.~Soricut, L.~Specia, and A.~Tamchyna, ``Findings of the 2014 workshop on statistical machine translation,'' in \emph{Proceedings of the Ninth Workshop on Statistical Machine Translation}.\hskip 1em plus 0.5em minus 0.4em\relax Baltimore, Maryland, USA: Association for Computational Linguistics, Jun. 2014, pp. 12--58. [Online]. Available: \url{https://aclanthology.org/W14-3302}
\BIBentrySTDinterwordspacing

\bibitem{bhosale-etal-2020-language}
\BIBentryALTinterwordspacing
S.~Bhosale, K.~Yee, S.~Edunov, and M.~Auli, ``Language models not just for pre-training: Fast online neural noisy channel modeling,'' in \emph{Proceedings of the Fifth Conference on Machine Translation}.\hskip 1em plus 0.5em minus 0.4em\relax Online: Association for Computational Linguistics, Nov. 2020, pp. 584--593. [Online]. Available: \url{https://aclanthology.org/2020.wmt-1.69}
\BIBentrySTDinterwordspacing

\bibitem{devlin-etal-2019-bert}
\BIBentryALTinterwordspacing
J.~Devlin, M.-W. Chang, K.~Lee, and K.~Toutanova, ``{BERT}: Pre-training of deep bidirectional transformers for language understanding,'' in \emph{Proceedings of the 2019 Conference of the North {A}merican Chapter of the Association for Computational Linguistics: Human Language Technologies, Volume 1 (Long and Short Papers)}.\hskip 1em plus 0.5em minus 0.4em\relax Minneapolis, Minnesota: Association for Computational Linguistics, Jun. 2019, pp. 4171--4186. [Online]. Available: \url{https://aclanthology.org/N19-1423}
\BIBentrySTDinterwordspacing

\bibitem{ott2019fairseq}
M.~Ott, S.~Edunov, A.~Baevski, A.~Fan, S.~Gross, N.~Ng, D.~Grangier, and M.~Auli, ``fairseq: A fast, extensible toolkit for sequence modeling,'' in \emph{Proceedings of NAACL-HLT 2019: Demonstrations}, 2019.

\bibitem{wolf-etal-2020-transformers}
\BIBentryALTinterwordspacing
T.~Wolf, L.~Debut, V.~Sanh, J.~Chaumond, C.~Delangue, A.~Moi, P.~Cistac, T.~Rault, R.~Louf, M.~Funtowicz, J.~Davison, S.~Shleifer, P.~von Platen, C.~Ma, Y.~Jernite, J.~Plu, C.~Xu, T.~Le~Scao, S.~Gugger, M.~Drame, Q.~Lhoest, and A.~Rush, ``Transformers: State-of-the-art natural language processing,'' in \emph{Proceedings of the 2020 Conference on Empirical Methods in Natural Language Processing: System Demonstrations}.\hskip 1em plus 0.5em minus 0.4em\relax Online: Association for Computational Linguistics, Oct. 2020, pp. 38--45. [Online]. Available: \url{https://aclanthology.org/2020.emnlp-demos.6}
\BIBentrySTDinterwordspacing

\bibitem{papineni-etal-2002-bleu}
\BIBentryALTinterwordspacing
K.~Papineni, S.~Roukos, T.~Ward, and W.-J. Zhu, ``{B}leu: a method for automatic evaluation of machine translation,'' in \emph{Proceedings of the 40th Annual Meeting of the Association for Computational Linguistics}.\hskip 1em plus 0.5em minus 0.4em\relax Philadelphia, Pennsylvania, USA: Association for Computational Linguistics, Jul. 2002, pp. 311--318. [Online]. Available: \url{https://aclanthology.org/P02-1040}
\BIBentrySTDinterwordspacing

\bibitem{heafield-2011-kenlm}
\BIBentryALTinterwordspacing
K.~Heafield, ``{K}en{LM}: Faster and smaller language model queries,'' in \emph{Proceedings of the Sixth Workshop on Statistical Machine Translation}.\hskip 1em plus 0.5em minus 0.4em\relax Edinburgh, Scotland: Association for Computational Linguistics, Jul. 2011, pp. 187--197. [Online]. Available: \url{https://aclanthology.org/W11-2123}
\BIBentrySTDinterwordspacing

\bibitem{guo-etal-2020-jointly}
\BIBentryALTinterwordspacing
J.~Guo, L.~Xu, and E.~Chen, ``Jointly masked sequence-to-sequence model for non-autoregressive neural machine translation,'' in \emph{Proceedings of the 58th Annual Meeting of the Association for Computational Linguistics}.\hskip 1em plus 0.5em minus 0.4em\relax Online: Association for Computational Linguistics, Jul. 2020, pp. 376--385. [Online]. Available: \url{https://aclanthology.org/2020.acl-main.36}
\BIBentrySTDinterwordspacing

\bibitem{huang2021improving}
X.~S. Huang, F.~Perez, and M.~Volkovs, ``Improving non-autoregressive translation models without distillation,'' in \emph{International Conference on Learning Representations}, 2021.

\bibitem{qian-etal-2021-glancing}
\BIBentryALTinterwordspacing
L.~Qian, H.~Zhou, Y.~Bao, M.~Wang, L.~Qiu, W.~Zhang, Y.~Yu, and L.~Li, ``Glancing transformer for non-autoregressive neural machine translation,'' in \emph{Proceedings of the 59th Annual Meeting of the Association for Computational Linguistics and the 11th International Joint Conference on Natural Language Processing (Volume 1: Long Papers)}.\hskip 1em plus 0.5em minus 0.4em\relax Online: Association for Computational Linguistics, Aug. 2021, pp. 1993--2003. [Online]. Available: \url{https://aclanthology.org/2021.acl-long.155}
\BIBentrySTDinterwordspacing

\bibitem{Li2020LAVANA}
X.~Li, Y.~Meng, A.~Yuan, F.~Wu, and J.~Li, ``Lava nat: A non-autoregressive translation model with look-around decoding and vocabulary attention,'' \emph{ArXiv}, vol. abs/2002.03084, 2020.

\bibitem{zhu2020incorporatingbertneuralmachine}
\BIBentryALTinterwordspacing
J.~Zhu, Y.~Xia, L.~Wu, D.~He, T.~Qin, W.~Zhou, H.~Li, and T.-Y. Liu, ``Incorporating bert into neural machine translation,'' 2020. [Online]. Available: \url{https://arxiv.org/abs/2002.06823}
\BIBentrySTDinterwordspacing

\bibitem{DBLP:journals/corr/abs-2001-05136}
\BIBentryALTinterwordspacing
J.~Kasai, J.~Cross, M.~Ghazvininejad, and J.~Gu, ``Parallel machine translation with disentangled context transformer,'' \emph{CoRR}, vol. abs/2001.05136, 2020. [Online]. Available: \url{https://arxiv.org/abs/2001.05136}
\BIBentrySTDinterwordspacing

\bibitem{geng-etal-2021-learning}
\BIBentryALTinterwordspacing
X.~Geng, X.~Feng, and B.~Qin, ``Learning to rewrite for non-autoregressive neural machine translation,'' in \emph{Proceedings of the 2021 Conference on Empirical Methods in Natural Language Processing}.\hskip 1em plus 0.5em minus 0.4em\relax Online and Punta Cana, Dominican Republic: Association for Computational Linguistics, Nov. 2021, pp. 3297--3308. [Online]. Available: \url{https://aclanthology.org/2021.emnlp-main.265}
\BIBentrySTDinterwordspacing

\bibitem{huang2022directed}
F.~Huang, H.~Zhou, Y.~Liu, H.~Li, and M.~Huang, ``Directed acyclic transformer for non-autoregressive machine translation,'' 2022.

\bibitem{Wang_Tian_He_Qin_Zhai_Liu_2019}
\BIBentryALTinterwordspacing
Y.~Wang, F.~Tian, D.~He, T.~Qin, C.~Zhai, and T.-Y. Liu, ``Non-autoregressive machine translation with auxiliary regularization,'' \emph{Proceedings of the AAAI Conference on Artificial Intelligence}, vol.~33, no.~01, pp. 5377--5384, Jul. 2019. [Online]. Available: \url{https://ojs.aaai.org/index.php/AAAI/article/view/4476}
\BIBentrySTDinterwordspacing

\bibitem{NEURIPS2019_74563ba2}
\BIBentryALTinterwordspacing
Z.~Sun, Z.~Li, H.~Wang, D.~He, Z.~Lin, and Z.~Deng, ``Fast structured decoding for sequence models,'' in \emph{Advances in Neural Information Processing Systems}, H.~Wallach, H.~Larochelle, A.~Beygelzimer, F.~d\textquotesingle Alch\'{e}-Buc, E.~Fox, and R.~Garnett, Eds., vol.~32.\hskip 1em plus 0.5em minus 0.4em\relax Curran Associates, Inc., 2019. [Online]. Available: \url{https://proceedings.neurips.cc/paper/2019/file/74563ba21a90da13dacf2a73e3ddefa7-Paper.pdf}
\BIBentrySTDinterwordspacing

\bibitem{song-etal-2021-alignart}
\BIBentryALTinterwordspacing
J.~Song, S.~Kim, and S.~Yoon, ``{A}lig{NART}: Non-autoregressive neural machine translation by jointly learning to estimate alignment and translate,'' in \emph{Proceedings of the 2021 Conference on Empirical Methods in Natural Language Processing}.\hskip 1em plus 0.5em minus 0.4em\relax Online and Punta Cana, Dominican Republic: Association for Computational Linguistics, Nov. 2021, pp. 1--14. [Online]. Available: \url{https://aclanthology.org/2021.emnlp-main.1}
\BIBentrySTDinterwordspacing

\bibitem{gu2019levenshteintransformer}
\BIBentryALTinterwordspacing
J.~Gu, C.~Wang, and J.~Zhao, ``Levenshtein transformer,'' 2019. [Online]. Available: \url{https://arxiv.org/abs/1905.11006}
\BIBentrySTDinterwordspacing

\bibitem{keles2022computational}
F.~D. Keles, P.~M. Wijewardena, and C.~Hegde, ``On the computational complexity of self-attention,'' 2022.

\bibitem{Jonker1987ASA}
\BIBentryALTinterwordspacing
R.~Jonker and T.~Volgenant, ``A shortest augmenting path algorithm for dense and sparse linear assignment problems,'' \emph{Computing}, vol.~38, pp. 325--340, 1987. [Online]. Available: \url{https://api.semanticscholar.org/CorpusID:7806079}
\BIBentrySTDinterwordspacing

\bibitem{DBLP:journals/corr/abs-1910-10683}
\BIBentryALTinterwordspacing
C.~Raffel, N.~Shazeer, A.~Roberts, K.~Lee, S.~Narang, M.~Matena, Y.~Zhou, W.~Li, and P.~J. Liu, ``Exploring the limits of transfer learning with a unified text-to-text transformer,'' \emph{CoRR}, vol. abs/1910.10683, 2019. [Online]. Available: \url{http://arxiv.org/abs/1910.10683}
\BIBentrySTDinterwordspacing

\bibitem{martins2020sparse}
A.~F.~T. Martins, A.~Farinhas, M.~Treviso, V.~Niculae, P.~M.~Q. Aguiar, and M.~A.~T. Figueiredo, ``Sparse and continuous attention mechanisms,'' 2020.

\bibitem{yang-etal-2022-textpruner}
\BIBentryALTinterwordspacing
Z.~Yang, Y.~Cui, and Z.~Chen, ``{T}ext{P}runer: A model pruning toolkit for pre-trained language models,'' in \emph{Proceedings of the 60th Annual Meeting of the Association for Computational Linguistics: System Demonstrations}.\hskip 1em plus 0.5em minus 0.4em\relax Dublin, Ireland: Association for Computational Linguistics, May 2022, pp. 35--43. [Online]. Available: \url{https://aclanthology.org/2022.acl-demo.4}
\BIBentrySTDinterwordspacing

\end{thebibliography}

%

\vfill
\end{document}